\documentclass{article}
\usepackage{newunicodechar}
\newunicodechar{≈}{\approx}

\usepackage[authoryear, sort&compress, round]{natbib}
\usepackage[preprint,logo]{agibot_enerverse} 

\usepackage[utf8]{inputenc} 
\usepackage[T1]{fontenc}    
\usepackage{hyperref}       
\usepackage{url}            
\usepackage{booktabs}       
\usepackage{amsfonts}       
\usepackage{nicefrac}       
\usepackage{microtype}      
\usepackage{xcolor}   
\usepackage{amsmath}
\usepackage{graphicx}
\usepackage{caption}
\usepackage{subcaption}
\usepackage{wrapfig}
\usepackage{pifont}
\usepackage{colortbl} 
\usepackage{multirow} 
\usepackage[table]{xcolor}
\usepackage{float}
\usepackage[absolute,overlay]{textpos}

\definecolor{green}{RGB}{0,150,10}
\definecolor{blue}{RGB}{0,148,181}
\definecolor{orange}{RGB}{194,153,107}

\definecolor{grey}{HTML}{999999}
\definecolor{lightblue}{HTML}{B0C4DE}
\definecolor{purple}{HTML}{E3BBED}
\definecolor{orange}{HTML}{ffdab9}
\definecolor{cadetblue}{HTML}{5F9EA0}
\definecolor{darksalmon}{rgb}{0.91, 0.59, 0.48}
\definecolor{forestgreen}{rgb}{0.13, 0.55, 0.13}
\definecolor{BlueGreen}{rgb}{0.0, 0.55, 0.55}
\definecolor{RedOrange}{rgb}{1.0, 0.27, 0.0}

\usepackage{xspace}

\author{
\parbox{0.96\textwidth}{\centering
{\small\bfseries
Boxiang Qiu$^{1,2,*}$ \quad
Liliang Chen$^{1,*,\dagger}$ \quad
Yue Liao$^{3}$ \quad
Nan Wang$^{1}$ \quad
Lintao Wang$^{1}$
\\[0.25em]
Jiayi Luo$^{1,2}$ \quad
Wenzhi Zhao$^{1,4}$ \quad
Shengcong Chen$^{1}$ \quad
Di Chen$^{1}$ \quad
Ye Li$^{4}$
\\[0.25em]
Chen Gao$^{2,3}$ \quad
Shuicheng Yan$^{3,\diamond}$ \quad
Si Liu$^{2,\diamond}$ \quad
Maoqing Yao$^{1,\diamond}$ \quad
Guanghui Ren$^{1,\dagger,\diamond}$
}
\\[0.8em]
{\normalsize $^{1}$ AgiBot \quad $^{2}$ BUAA \quad $^{3}$ LV-NUS Lab \quad $^{4}$ TJU}
\\[0.4em]
{\normalsize \url{https://ge-sim-v2.github.io/}}
}
}

\title{GE-Sim 2.0: A Roadmap Towards Comprehensive Closed-loop Video World Simulators for Robotic Manipulation}

\makeatletter
\renewcommand{\@noticestring}{%
  $^*$ Equal Contribution. \quad
  $^\dagger$ Project Leader. \quad
  $^\diamond$ Corresponding Author.%
}
\makeatother

\begin{document}

\maketitle

\begin{abstract}

We introduce \textbf{GE-Sim 2.0 (Genie Envisioner World Simulator 2.0)}, 
a closed-loop video world simulator for robotic manipulation. Building on 
the action-conditioned video generation framework of Genie Envisioner, 
GE-Sim 2.0 is re-trained on thousands of hours of real-world robot data spanning teleoperation, contact-rich interaction, and on-robot policy deployment, substantially improving action-following fidelity and 
trajectory coverage. On top of this foundation, three new modules close the loop from video simulation to policy learning: a \emph{state expert} that decodes proprioceptive state from video latents to support next-chunk prediction by downstream policy models; a \emph{world judge} that scores generated rollouts against task instructions, yielding machine-verifiable success signals and rewards in place of manual inspection; and an \emph{acceleration framework} that delivers a 
25-frame rollout in 2.3 seconds on a single H100, with up to 
$4\times$ frame skipping at inference for long-horizon evaluation. GE-Sim 2.0 tops the public WorldArena leaderboard at only 2B parameters, outperforming both dedicated robotic world models and closed-source general video generators, and policies trained against its rollouts and rewards translate into measurable real-world gains, establishing GE-Sim 2.0 as a practical platform for scalable evaluation and 
closed-loop learning of manipulation policies.

\end{abstract}

\begin{figure}[ht]
    \begin{center}
\centerline{\includegraphics[width=1\linewidth]{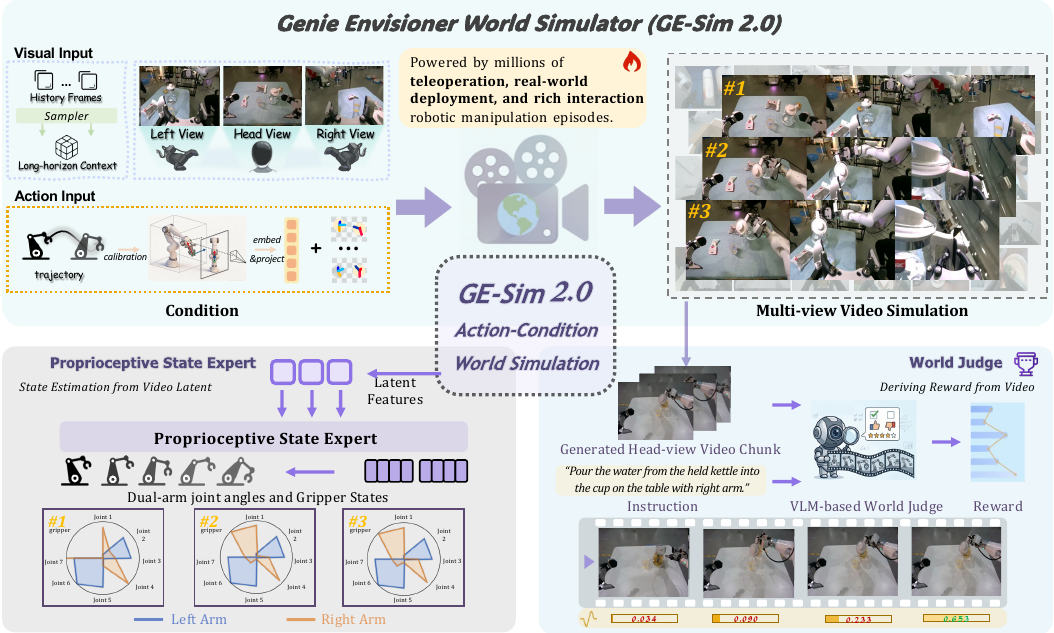}}
    \caption{\textbf{Overview of GE-Sim 2.0.} 
GE-Sim 2.0 is a closed-loop video world simulator for robotic 
manipulation, trained on millions of real-world episodes spanning 
teleoperation, on-robot policy deployment, and rich object interaction. 
Given long-horizon multi-view history frames and an action trajectory 
embedded from end-effector calibration, the model generates 
action-conditioned multi-view rollouts of the robot executing the 
specified behavior \emph{(top)}. Two complementary modules close the 
loop on top of visual simulation \emph{(bottom)}: a \textbf{proprioceptive 
state expert} decodes dual-arm joint angles and gripper states from 
video latents, supplying downstream policy models with the state 
information needed for next-chunk prediction; and a VLM-based 
\textbf{world judge} scores generated rollouts against the task 
instruction.}
        \label{fig:teaser}
    \end{center}
\end{figure}

\section{Introduction}

Robot learning is entering a scaling era. Larger models, internet-scale demonstrations, and increasingly capable vision-language-action (VLA) policies~\citep{brohan2022rt,brohan2024rt,driess2023palm,zhu2023vima,team2024octo,black2024pi0,intelligence2025pi05,kim2024openvla,generalist2025gen0,generalist2026gen1} are pushing manipulation beyond rigid-body pick-and-place toward long-horizon, contact-rich, and deformable-object tasks. Yet as policies scale, \emph{evaluation} has become the bottleneck: real-robot benchmarking is slow and hard to reproduce, while existing robotic benchmarks and simulators~\citep{james2020rlbench,mees2022calvin,liu2023libero,todorov2012mujoco,Makoviychuk2021IsaacGH,Xiang2020SAPIENAS,Gu2023ManiSkill2AU,Lin2020SoftGymBD} still struggle with contact dynamics, deformable objects, fine-grained visual appearance, and even the robot's own actuation, where effects such as harmonic-drive compliance are routinely abstracted away. The gap between what we can train and what we can reliably evaluate keeps widening.

Recent progress in generative video modeling~\citep{Singer2022MakeAVideoTG,Villegas2022PhenakiVL,Blattmann2023AlignYL,BarTal2024LumiereAS,openai2024sora} offers a different path. Trained on web-scale video, modern generators can synthesize photorealistic footage across a wide diversity of scenes, objects, and interactions that handcrafted simulators cannot easily reproduce. This motivates a new paradigm, a \emph{neural world simulator} for manipulation: given an initial observation and an action trajectory from a policy, human, or teleoperation, the model rolls out a video of the robot executing that behavior in a learned visual world. By replacing hand-built physics and rendering with a data-driven generative process, such a simulator promises to cover the long tail of real-world appearances and interactions that classical engines miss, opening a path toward both scalable evaluation and closed-loop policy learning, including reinforcement learning, of 
modern manipulation policies.

A growing line of work~\citep{ho20251x,wang2026interactive,jiang2025enerverseac,liao2025genie,guo2025ctrl,zhu2024irasim,nvidia2025cosmos,gao2026dreamdojo} has begun to explore video-based world simulators for manipulation, including our own GE-Sim~\citep{liao2025genie}. These efforts share a common recipe: re-purpose a pretrained \emph{text-image-to-video} (TI2V) generator into an \emph{action-image-to-video} simulator by replacing the text condition with an action condition, and most technical effort to date has focused on \emph{how} to inject this action signal so that the generated video faithfully follows a given trajectory. While such systems already produce visually plausible action-conditioned rollouts, their fidelity on deformable objects and on out-of-distribution or failure trajectories remains limited. More fundamentally, even a perfectly photorealistic action-following video does not, by itself, constitute a simulator that a policy can close the loop on. Three gaps stand out: (i) existing simulators predict only future visual states, leaving unmodeled the proprioceptive state that modern policy models require, with commanded actions used as a noisy proxy that drifts from the arm's actual motion; (ii) they render rollouts but do not score them, withholding the verifiable signals required for scalable evaluation and reward-driven learning; and (iii) their rendering throughput is far below what chunk-wise, parallel rollout across many tasks and seeds demands. Together, these three gaps mark the path from today's action-conditioned video models toward what we call a \textbf{closed-loop world simulator} for manipulation.

To close these gaps, we introduce \textbf{GE-Sim 2.0} (Genie Envisioner World Simulator 2.0), illustrated in Figure~\ref{fig:teaser}. GE-Sim 2.0 retains the action-conditioned video generation backbone of GE-Sim~\citep{liao2025genie}, and is re-trained on thousands of hours of real-world robot data, combining large-scale teleoperation episodes, contact-rich arm-object interaction sequences, and rollouts collected during policy deployment on physical robots. Conditioned on long-horizon multi-view history frames and an action trajectory, the model rolls out action-following multi-view videos of the robot executing the specified behavior. This data scale and diversity substantially improve action-following accuracy, the visual fidelity of object deformation and contact, and the coverage of diverse trajectories across successful executions, failure cases, and varied tasks and scenes. On top of this strengthened foundation, we introduce three modules that directly address the three gaps identified above. First, a \textbf{state expert} decodes proprioceptive state, namely dual-arm joint angles and gripper states, from video latents, providing downstream policy models with faithful state alongside visual observations for next-chunk prediction. Second, a VLM-based \textbf{world judge} scores generated rollouts against task instructions, turning simulator output into machine-verifiable success signals and rewards for both policy 
evaluation and reward-driven learning. Third, an \textbf{acceleration 
framework} improves rollout throughput without sacrificing fidelity, 
making chunk-wise, large-scale rollout tractable. Together, these 
components advance GE-Sim 2.0 from a visual simulator into a 
closed-loop, machine-verifiable platform for scalable manipulation 
policy training and evaluation.

We validate GE-Sim 2.0 along the design axes outlined above. As a foundation video simulator, it tops the public WorldArena leaderboard~\citep{shang2026worldarena} at only 2B parameters, outperforming dedicated robotic world models such as Ctrl-World~\citep{guo2025ctrl}, DreamDojo~\citep{gao2026dreamdojo}, GigaWorld~\citep{gigaworld2025}, and ABot~\citep{chen2026abot}, alongside closed-source general video generators including Sora~\citep{openai2024sora} and Veo~\citep{deepmind2025veo3}. At a finer granularity, per-task replay metrics (PSNR, SSIM, LPIPS, FID, FVD) across six representative manipulation tasks show that this advantage holds case by case, and closed-loop evaluation shows that policy outcomes inside the simulator agree with the real robot at both the aggregate and per-episode level. The state expert recovers proprioceptive state with high fidelity and improves downstream next-chunk prediction, while the world judge yields reward signals closely aligned with human judgment. Driven by the acceleration framework, GE-Sim 2.0 generates a 25-frame rollout in 2.3 seconds on a single H100, and a random-stride training scheme enables up to $4\times$ frame skipping at inference, extending the simulated horizon without measurable loss in evaluation consistency. Crucially, policies trained against GE-Sim 2.0's rollouts and rewards translate into measurable real-world gains, lifting it from a passive video generator into an active driver of policy learning.

\section{Preliminaries}
\label{sec:prelim}

GE-Sim 2.0 builds on the Genie Envisioner platform~\citep{liao2025genie}, 
in particular its world foundation model \emph{GE-Base} and its 
action-conditioned simulator \emph{GE-Sim}. We briefly review both here 
to fix notation; readers familiar with Genie Envisioner  may skip 
this section.

\subsection{GE-Base: Multi-View Video World Foundation Model}
\label{sec:prelim_gebase}

GE-Base formulates robotic world modeling as a multi-view 
text-and-image-to-video generation problem. Given a language instruction 
$q$ and an initial multi-view observation, the model autoregressively 
predicts future video chunks that capture how the scene evolves under 
the instruction.

\textit{Autoregressive chunk-wise generation.}\quad
Let $\mathcal{V} = \{h, l, r\}$ denote the set of onboard cameras 
(head, left wrist, right wrist), and let $x_t^{i}$ denote the frame 
from view $i \in \mathcal{V}$ at time $t$. At autoregressive step $t$, 
the world model $\mathcal{W}$ predicts the next chunk of $N$ multi-view 
frames $\mathbf{x}^{t}_{1:N}$ conditioned on the initial observation 
$\mathbf{x}_0$, a long-term \emph{sparse memory} $\mathbf{m}_{0:t-1}$, 
and the encoded instruction $\mathcal{T}(q)$:
\begin{equation}
\mathbf{x}^{t}_{1:N} \;=\; 
\mathcal{W}\!\left(\mathbf{x}_0,\, \mathbf{m}_{0:t-1},\, \mathcal{T}(q)\right),
\label{eq:gebase_ar}
\end{equation}
where $\mathbf{m}_{0:t-1}$ is constructed by sparsely sampling keyframes 
from previously generated chunks $\{\mathbf{x}^{k}_{1:N}\}_{k=0}^{t-1}$, 
and $\mathcal{T}(\cdot)$ is a frozen T5 text encoder. This sparse memory 
mechanism extends the temporal context of the model far beyond the 
current chunk while keeping the input length tractable.

\textit{Multi-view encoding.}\quad
Each per-view input is processed independently by a shared video 
encoder $\mathcal{E}$, producing initial and memory tokens 
$v_0^{i} = \mathcal{E}(x_0^{i})$ and 
$v_m^{i} = \mathcal{E}(m_{0:t-1}^{i})$. Each token is enriched with a 
3D rotary positional embedding and a learnable view embedding $e^{i}_{\text{view}}$:
\begin{equation}
\tilde{v}^{i} \;=\; \mathrm{RoPE}(t, h, w) + v^{i} + e^{i}_{\text{view}}.
\end{equation}
Together with a view-specific noise map $z^{i}$, the per-view input 
sequence is $u^{i} = \big[\tilde{v}_0^{i} \,\|\, \tilde{v}_m^{i} \,\|\, z^{i}\big]$. 
Tokens from all views are concatenated and processed by a video 
diffusion transformer (DiT). To enforce cross-view consistency, a subset 
of DiT blocks performs \emph{cross-view attention} over the merged 
multi-view sequence, while the remaining blocks treat views independently 
for efficiency.

\textit{Backbone and training.}\quad
We instantiate $\mathcal{W}$ with the Cosmos-Predict2-2B-Video2World 
DiT~\citep{nvidia2025cosmos}, which provides strong visual priors for 
high-fidelity simulation. Training follows a latent flow-matching 
objective: given the VAE latent $l$ of the target chunk and a noisy 
latent $\tilde{l} = (1 - \sigma_\tau)\,l + \sigma_\tau\,\epsilon$ with 
$\epsilon \sim \mathcal{N}(0, I)$, $\mathcal{W}$ predicts the denoising 
velocity $v_\theta$, supervised on future frames via a conditioning 
mask $M$:
\begin{equation}
\mathcal{L}_{\text{video}} \;=\; w(\tau)\,
\big\| \big(v_\theta - (\epsilon - l)\big) \odot (1 - M) \big\|_2^2.
\label{eq:gebase_loss}
\end{equation}

\subsection{Genie Envisioner World Simulator (GE-Sim)}
\label{sec:prelim_gesim}

GE-Sim repurposes GE-Base from a text-and-image-to-video generator into 
an \emph{action-conditioned} video simulator: instead of being driven 
by a language instruction, generation is driven by a low-level robot 
action trajectory, so that the synthesized video faithfully reflects 
how the robot would execute that trajectory in the scene.

\textit{Action representation.}\quad
For a dual-arm system, each control step is encoded as a 14-dimensional 
vector formed by concatenating the 7-D end-effector states of both arms:
\begin{equation}
a_i \;=\; \big[\,\underbrace{x_i, y_i, z_i, r_i, p_i, y_i, o_i}_{\text{left arm}},\;
\underbrace{x_i, y_i, z_i, r_i, p_i, y_i, o_i}_{\text{right arm}}\,\big] \in \mathbb{R}^{14},
\end{equation}
where $(x, y, z)$ is the end-effector position, $(r, p, y)$ its 
roll-pitch-yaw orientation, and $o$ the gripper openness. Over a 
$K$-step horizon, the full trajectory is denoted 
$\mathbf{A} = [a_1, \ldots, a_K] \in \mathbb{R}^{K \times 14}$.

\textit{Spatial action conditioning.}\quad
Bridging the low-level control signal $\mathbf{A}$ and the high-dimensional
latent space of $\mathcal{W}$ requires a spatially-aligned conditioning
signal that specifies both \emph{where} the end effector should appear
in the image plane and \emph{from which viewpoint} each frame is observed.
GE-Sim therefore couples a Pose2Image rendering with an explicit camera
raymap, both of which live in the pixel grid of the target view.

\textbf{Pose image.}
At each step $i$, the position $(x_i, y_i, z_i)$ is projected into
pixel coordinates via the calibrated camera intrinsics and extrinsics,
the orientation axes are projected as directional unit vectors, and the
gripper openness is rendered on a unit circle whose shading reflects
$o_i$. Distinct color encodings differentiate the two arms. This yields
a \emph{pose image} $P_i \in \mathbb{R}^{3 \times H \times W}$ that is
spatially aligned with the scene.

\textbf{Camera raymap.}
To make the multi-view camera geometry explicit, we additionally construct
a per-pixel \emph{raymap} $R_i \in \mathbb{R}^{6 \times H \times W}$ from
the intrinsics $K$ and the camera-to-world extrinsics $T_i$. For each
pixel $(u, v)$, we form the ray origin $\mathbf{o}_i \in \mathbb{R}^3$
(the camera center in world coordinates) and the unit-norm view direction
$\mathbf{d}_i \in \mathbb{R}^3$ obtained by back-projecting $(u, v)$
through $K$ and rotating by the camera-to-world rotation in $T_i$.
Stacking $\mathbf{o}_i$ and $\mathbf{d}_i$ along the channel dimension
gives a 6-channel raymap that exposes the camera pose at every pixel,
so the simulator does not have to infer viewpoint geometry from appearance
alone.

\textbf{Latent fusion.}
Both $P_i$ and $R_i$ are bilinearly downsampled to the latent spatial
resolution and concatenated with the noisy video latent $\tilde{z}_i$
along the channel dimension, so the simulator consumes a single fused
input
\begin{equation}
v_i \;=\; \big[\,\tilde{z}_i \,\Vert\, \mathrm{down}(P_i) \,\Vert\, \mathrm{down}(R_i)\,\big].
\label{eq:pose2img}
\end{equation}
This channel-wise fusion preserves spatial alignment between the action
condition, the camera geometry, and the latent video tokens at every
layer of the DiT backbone.

\textit{From TI2V to action-conditioned simulation.}\quad
With this conditioning mechanism in place, the role of the language
condition $\mathcal{T}(q)$ in Eq.~\eqref{eq:gebase_ar} is replaced by
the action trajectory $\mathbf{A}$, while the multi-view, autoregressive,
sparse-memory backbone of GE-Base is kept intact. The simulator thus 
predicts
\begin{equation}
\mathbf{x}^{t}_{1:N} \;=\; 
\mathcal{S}\!\left(\mathbf{x}_0,\, \mathbf{m}_{0:t-1},\, \mathbf{A}^{t}\right),
\label{eq:gesim}
\end{equation}
where $\mathbf{A}^{t}$ is the action sub-trajectory aligned with chunk 
$t$, and $\mathcal{S}$ denotes the action-conditioned simulator obtained 
by replacing $\mathcal{T}(q)$ with the hierarchical action conditioning 
described above. This formulation forms the foundation on which GE-Sim 2.0 
is built.
\section{Genie Envisioner World Simulator 2.0}
\label{sec:method}
In this section, we introduce \textbf{GE-Sim 2.0}, a comprehensive upgrade 
of GE-Sim. GE-Sim 2.0 advances a ``view-only'' video world simulator into 
a closed-loop world simulator that interacts precisely with a policy model 
and provides feedback on the interaction.
\subsection{Overview}
\label{sec:method_overview}

\begin{figure}[t]
\centering
\vspace*{-15pt}
\hspace{-20pt}
\includegraphics[width=\linewidth]{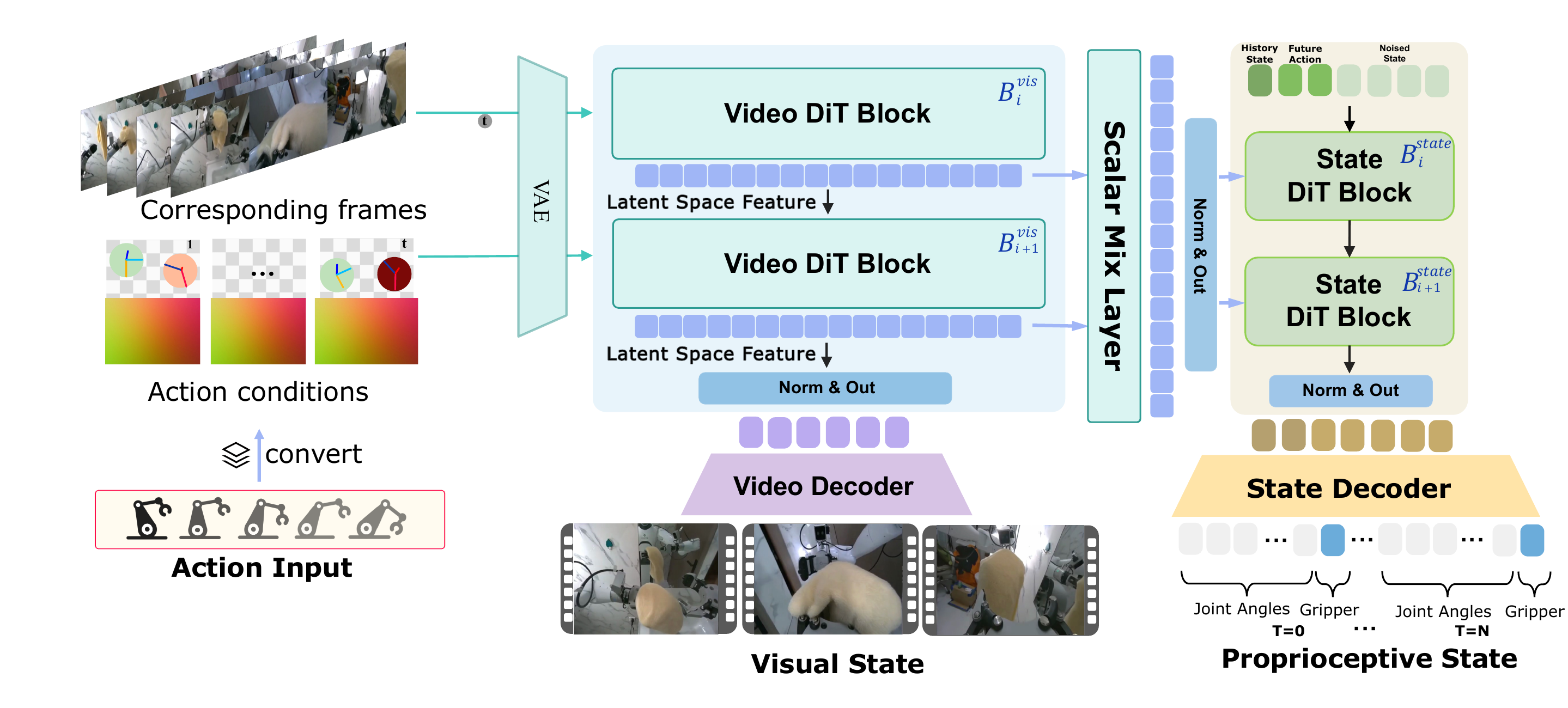}
\caption{\textbf{Vision expert and proprioceptive state expert overview.} The vision expert processes historical frames and action conditions to generate future visual states, which are then consumed by the proprioceptive state expert to predict the joint angles and gripper openness of both arms.}
\label{fig:video_state_dit}
\end{figure}

As shown in Figure~\ref{fig:video_state_dit}, at each autoregressive step, GE-Sim 2.0 takes as input an initial observation $\mathbf{x}_0$, a sparse memory $\mathbf{m}$, and an action sequence $\mathbf{A}^t$. In the canonical deployment scenario $\mathbf{A}^t$ is produced by a policy model under evaluation, but the simulator is agnostic to the source and equally accepts actions from teleoperation logs, motion planners, or hand-authored trajectories. Following the Pose2Image formulation in Section~\ref{sec:prelim_gesim}, every action is rendered into a visually aligned EE pose map using the camera intrinsics and extrinsics, so that the position, orientation, and openness of each gripper are explicitly encoded as pixel-level conditions. This single action representation is shared across all components of GE-Sim 2.0.

GE-Sim 2.0 is organized as two parallel experts that share the same conditioning. The \textbf{vision expert} is an action-conditioned diffusion transformer that follows the chunk-wise autoregressive and sparse-memory framework of Section~\ref{sec:prelim_gebase} and generates a future video chunk $\hat{\mathbf{x}}^t$, predicting what the robot would observe under the given action sequence and serving as the visual backbone of the simulator. The \textbf{proprioceptive state expert} runs in parallel with the vision expert and predicts the corresponding joint-space state sequence $\hat{\mathbf{s}}^t$ by consuming the intermediate features of the vision expert as visual context, recovering the proprioceptive observation that a real robot would return. When the policy and GE-Sim 2.0 are coupled into a closed-loop system, the two expert outputs $(\hat{\mathbf{x}}^t, \hat{\mathbf{s}}^t)$ are fed back to the policy as the input for its next chunk of action prediction, extending the simulator from pure visual rendering to a source of the full observation required for policy interaction, and forming an interaction loop consistent with the real robot.

On top of this loop, we attach a \textbf{world judge}, a vision-language reward model that scores the rollout generated by the world simulator frame by frame and outputs a machine-verifiable success signal $\hat{r}^t$. This signal serves both as an automated referee for policy evaluation and as sparse feedback for downstream reward-driven learning such as filtered BC and RL, giving GE-Sim 2.0 a built-in critic capability.

GE-Sim 2.0 is built in stages. We first train the vision expert on thousands of hours of real-world robot data (Section~\ref{sec:method_foundation}). We then freeze the vision expert and train the proprioceptive state expert so that it decodes joint-space state from the visual context of the vision expert (Section~\ref{sec:method_state_expert}). The world judge is trained independently as an external module (Section~\ref{sec:method_world_judge}). After the main training, we apply a distillation-based post-training acceleration to the world simulator (Section~\ref{sec:method_acceleration}), compressing multi-step diffusion into few-step inference. The following four sections describe these components in turn.

\subsection{Action-Conditioned Video Generation}
\label{sec:method_foundation}

The vision expert of GE-Sim 2.0 is an action-conditioned multi-view diffusion model and serves as the visual backbone of the simulator. It inherits the chunk-wise autoregressive generation, sparse memory, and multi-view diffusion transformer of GE-Base described in Section~\ref{sec:prelim_gebase}. On this basis, to turn the vision expert into a high-fidelity simulator suitable for closed-loop policy interaction, we describe its conditioning interface, action representation, and training.

\textbf{Conditioning interface.}\quad
The vision expert operates in the latent space of a video VAE, and all of its conditions enter the network through channel-wise concatenation. At each autoregressive step, the network input is
\begin{equation}
\mathbf{z}_{\text{cond}} = \big[\, \mathbf{z}_{\text{noisy}};\ \mathbf{R}_{\text{ray}};\ \mathbf{M}_{\text{pose}};\ \mathbf{m}_{\text{cond}} \,\big],
\label{eq:cond_interface}
\end{equation}
where $\mathbf{z}_{\text{noisy}}$ is the 16-channel noisy video latent, $\mathbf{R}_{\text{ray}}$ is a 6-channel per-pixel ray map, $\mathbf{M}_{\text{pose}}$ is a 3-channel EE pose map, and $\mathbf{m}_{\text{cond}}$ is a 1-channel binary mask distinguishing memory frames from frames to be predicted. The ray map and the EE pose map jointly form the action representation introduced next. All visual inputs are normalized from $[0,1]$ to $[-1,1]$ before encoding, and both conditioning maps follow the same $[-1,1]$ convention, so that video and conditioning channels are numerically aligned and diffusion training is not destabilized by scale mismatch.

\textbf{Action representation.}\quad
GE-Sim 2.0 conveys the dual-arm action to the vision expert through two visually aligned channels, the ray map $\mathbf{R}_{\text{ray}}$ and the EE pose map $\mathbf{M}_{\text{pose}}$. Together they encode \emph{what the camera sees of the action}: the ray map captures how the viewpoint moves with the robot, while the EE pose map captures how the end effectors move within that view. Decoupling the two factors is essential in our setup, because the head and wrist cameras are themselves mounted on the moving robot and the wrist views in particular shift substantially with the arm.

\emph{Ray map.}\quad
For each pixel, we build a ray in the world frame from the per-frame camera intrinsics and extrinsics, represented jointly by its origin and unit direction (six channels in total). As the cameras move with the robot, the ray map changes accordingly and gives the vision expert an explicit camera-geometry prior, allowing it to separate appearance changes caused by viewpoint motion from those caused by object motion in the scene. This is especially important for the wrist cameras, where the end effector is nearly static relative to the camera and the ray map carries most of the kinematic signal of the arm.

\emph{EE pose map.}\quad
Following the Pose2Image formulation of GE-Sim in Section~\ref{sec:prelim_gesim}, we render the future end-effector trajectory of both arms into the image space of each camera view, producing a 3-channel EE pose map spatially aligned with the scene. For each timestep and each view, the rendering proceeds in three steps. \emph{(i) Pose projection.} We express the gripper pose of each arm as a $4\times 4$ transformation in the world frame, transform it into the camera frame through a fixed wrist-to-EE correction, and project the EE origin together with one keypoint along each of the three coordinate axes onto the image plane, obtaining the pixel coordinates of the EE position and its orientation. \emph{(ii) Depth-aware rendering.} On the canvas, we draw a filled circle at the EE origin whose radius decreases monotonically with the distance from the camera to the EE,
\begin{equation}
r = \mathrm{clamp}\!\left(1 - \frac{\|\mathbf{x}_{\text{EE}} - \mathbf{x}_{\text{cam}}\| - d_{\min}}{d_{\max} - d_{\min}},\ 0,\ 1\right)\cdot r_{\max},
\label{eq:depth_radius}
\end{equation}
so that the closer the EE is to the camera, the larger the circle. The EE orientation is drawn as colored segments connecting the origin to the three projected axis keypoints, with the left and right arms using distinct color schemes. \emph{(iii) Gripper-openness encoding.} The fill color of the circle encodes the gripper openness through a continuous colormap: as the gripper goes from closed to open, the color goes from dark to light, with left and right arms using different color families. This jointly encodes the gripper state, which is binary in nature yet carries a continuous degree, in a single, continuous, and stable channel. The same renderer is used at training and inference time to keep the two input distributions aligned. The EE pose map serves as the unified pose-level action condition shared by the vision expert, the proprioceptive state expert, and the world judge.

\textbf{Training.}\quad
We train the vision expert on thousands of hours of real-world robot data. The training data spans teleoperation recordings, on-robot policy deployment rollouts, and a large number of object-interaction trajectories, and covers both successful and failed trajectories, improving the model's ability to simulate diverse actions and complex interactions. Training follows the flow-matching objective $\mathcal{L}_{\text{video}}$ of GE-Base in Section~\ref{sec:prelim_gebase}, with the loss computed only on frames to be predicted while memory frames are masked out by $\mathbf{m}_{\text{cond}}$. During closed-loop inference, the vision expert rolls out chunk by chunk and the memory frames of each chunk are sparsely sampled from previously generated content. Since these memory frames come from the vision expert's own generations, they inevitably contain generation artifacts and accumulated errors from continuous rollout, and are therefore not drawn from the same distribution as the clean memory frames used during training. To mitigate this distribution shift, we introduce a randomized error-simulation mechanism for memory frames during training, approximating the error patterns of self-generated memory frames through pre-encoding perturbation, local degradation, and multi-view synchronized appearance variation.

\subsection{Proprioceptive State Expert}
\label{sec:method_state_expert}

Closed-loop interaction requires not only future visual observations but also the robot's \emph{proprioceptive state}, which is one of the key inputs that modern policy models rely on for next-chunk action prediction. The vision expert generates only future video and does not directly expose the proprioceptive state of the arms in the rendered frames. To fill this gap, we introduce the \textbf{proprioceptive state expert}, a lightweight transformer branch that runs in parallel with the vision expert and predicts the proprioceptive state sequence over the corresponding time interval from the visual context produced by the vision expert.

\textbf{Proprioceptive state representation.}\quad
The proprioceptive state expert represents and regresses the proprioceptive state in joint space. The proprioceptive state at each frame is a 16-dimensional vector formed by concatenating the joint angles and gripper openness of both arms,
\begin{equation}
\mathbf{s}_t = \big[\, \boldsymbol{\theta}^{L}_t,\ g^{L}_t,\ \boldsymbol{\theta}^{R}_t,\ g^{R}_t \,\big] \in \mathbb{R}^{16},
\label{eq:state_repr}
\end{equation}
where $\boldsymbol{\theta}^{L}_t, \boldsymbol{\theta}^{R}_t \in \mathbb{R}^{7}$ are the seven joint angles of the left and right arms, and $g^{L}_t, g^{R}_t$ are the corresponding gripper openness values.
The gripper openness is linearly normalized to $[0,1]$, consistent with the gripper coordinate convention used by the EE pose map in Section~\ref{sec:method_foundation}. Besides the current interval to be predicted, which spans $T_{\text{fut}}$ future frames, the input sequence of the proprioceptive state expert also contains the proprioceptive state of $n_{\text{prev}}$ history frames together with the aligned history and future actions, giving $2 n_{\text{prev}} + 2 T_{\text{fut}}$ tokens in total, of which only the future proprioceptive state tokens are noised. This joint-space proprioceptive state complements the visual prediction, and together they form the full observation required for downstream closed-loop policy interaction.

\textbf{Visual context from the vision expert.}\quad
The visual information that the proprioceptive state expert needs for predicting the proprioceptive state comes from the vision expert, but the proprioceptive state expert is not aligned with the vision expert layer by layer. The vision expert first runs a complete forward pass through all of its $L$ transformer blocks, and we aggregate the output $\mathbf{h}^{\text{video}}_l$ of each layer into a \emph{single fused visual feature} through a set of learnable scalar weights $\alpha_l$ (initialized to one), followed by a single LayerNorm,
\begin{equation}
\mathbf{H}_{\text{fuse}} = \mathrm{LayerNorm}\!\left( \sum_{l=1}^{L} \alpha_l\, \mathbf{h}^{\text{video}}_l \right), \qquad \alpha_l \in \mathbb{R}.
\label{eq:hfuse}
\end{equation}
All $L$ blocks of the proprioceptive state expert share the same $\mathbf{H}_{\text{fuse}}$ as the key and value of their cross-attention. Under the multi-view setting, $\mathbf{H}_{\text{fuse}}$ is rearranged into $\mathbf{H}_{\text{fuse}} \in \mathbb{R}^{B \times (V L_{\text{tok}}) \times d}$, concatenating the tokens of all $V$ camera views along the sequence dimension, so that the proprioceptive state expert attends to the visual features of all views at once.

\textbf{Architecture.}\quad
The proprioceptive state expert is a stack of $L$ lightweight transformer blocks whose hidden dimension is much smaller than that of the vision expert. Each block consists of three parts: a self-attention under rotary positional embedding along the proprioceptive state time axis, a cross-attention to the vision-expert visual feature $\mathbf{H}_{\text{fuse}}$, and a feed-forward layer; the diffusion timestep is injected through AdaLN-single, whose scale-shift parameters use the same diffusion timestep as the vision expert. The proprioceptive state sequence is first projected into the hidden space of the proprioceptive state expert by a linear layer and then projected back to the $d_s$-dimensional state space by another linear layer at the end, where $d_s$ denotes the dimension of the proprioceptive state representation.

\textbf{Training and loss.}\quad
The proprioceptive state expert is trained on top of a frozen vision expert: we use the trained vision expert as the provider of visual context and update only the proprioceptive state expert. Training adopts the same flow-matching objective as the vision expert, applied to the proprioceptive state sequence. Given a clean proprioceptive state sequence $\mathbf{s}_0$ and noise $\boldsymbol{\epsilon}$, we construct a noised sequence $\mathbf{s}_\tau$, and the proprioceptive state expert $v_\phi$ predicts the denoising velocity, with the loss computed only on the future proprioceptive state,
\begin{equation}
\mathcal{L}_{\text{proprio state}} = \mathbb{E}_{\mathbf{s}_0,\boldsymbol{\epsilon},\tau}\big\| v_\phi(\mathbf{s}_\tau, \tau, \mathbf{H}_{\text{fuse}}) - (\boldsymbol{\epsilon} - \mathbf{s}_0) \big\|_2^2.
\label{eq:state_loss}
\end{equation}

\textbf{History-state augmentation.}\quad
The core input of the proprioceptive state expert is the history proprioceptive state, the history actions, and the future actions, and only the future proprioceptive state is noised. In real data, however, the history proprioceptive state is almost always a noise-free exact reading, whereas during closed-loop inference part of the history proprioceptive state comes from the proprioceptive state expert's own prediction at the previous chunk and carries error. Without intervention, the proprioceptive state expert would overly rely on a ``perfect history'' and be insufficiently robust under the closed loop.
We therefore apply two targeted perturbations to the history segment during training: temporal index perturbation, which simulates short-range misalignment between the policy and the vision expert, and history-trajectory resampling, which mimics slight temporal stretching or compression under asynchronous closed-loop execution. Detailed perturbation probabilities and parameter settings are provided in the appendix.
\begin{equation}
\mathbf{s}_{\text{hist}} \leftarrow \mathrm{Upsample}\big(\mathrm{Downsample}(\mathbf{s}_{\text{hist}},\, n_{\text{prev}}-1),\, n_{\text{prev}}\big),
\label{eq:hist_resample}
\end{equation}
which is equivalent to injecting a low-pass distortion that preserves the long-term trend while removing single-frame high-frequency detail, simulating the case where the history is fed back by the network rather than taken from ground truth.

\subsection{World Judge}
\label{sec:method_world_judge}

A central goal of GE-Sim 2.0 is to support large-scale policy evaluation. The world simulator generates rollouts but does not judge how good a rollout is. To make the simulator truly serve evaluation and reward-driven learning, we need a component that turns generated results into machine-verifiable signals. To this end, we introduce the \textbf{world judge}, a vision-language reward model that scores the rollout generated by GE-Sim 2.0 frame by frame during closed-loop inference and outputs a task-completion success signal. The design of the world judge builds on Robometer~\citep{liang2026robometer}, a general-purpose vision-language reward model; we adapt its formulation to our closed-loop setting and train a single success objective on real-robot data.

\textbf{Backbone and per-frame representation.}\quad
The world judge uses a vision-language model as its backbone. We freeze its vision encoder and train only the language model and the downstream prediction head. Each frame of the rollout is fed in individually as a separate image, without averaging over the time axis, so as to preserve per-frame discriminative granularity. After each frame image, we append a dedicated per-frame token and take the hidden state of this token as the representation $\mathbf{f}_i$ of that frame. A rollout of length $T$ is thus encoded into a set of per-frame representations $\{\mathbf{f}_i\}_{i=1}^{T}$. The textual condition of the world judge is not the full task instruction but the sub-task caption matched to the current chunk, so that the judgment is conditioned on the specific sub-task that the chunk is expected to accomplish.

\textbf{Success head and supervision.}\quad
While Robometer is trained with a dual progress-and-preference objective, our world judge deliberately focuses on a single per-frame success prediction. This design is motivated by two considerations. First, a sparse success signal is sufficient for closed-loop evaluation and downstream reinforcement learning, and has been widely adopted as the primary metric or reward signal in robotic manipulation benchmarks and offline RL settings~\citep{Andrychowicz2017HindsightER,Lu2025VLARLTM}. Second, progress supervision is substantially less reliable in our setting. In real robot data collection, trajectories often contain conscious or unconscious error recovery behaviors, such as detours, retries, and temporary regressions before eventual success. Under such non-monotonic executions, a scalar progress label becomes noisy and may no longer correspond to the true task completion process. We therefore discard progress prediction and formulate world judging as a class-balanced per-frame binary success classification problem.

The world judge attaches a success head, a lightweight MLP, on top of each per-frame representation, producing a success logit $\hat{s}_i = \mathrm{head}(\mathbf{f}_i)$ for that frame. The supervision comes from the manual success annotation of each trajectory: every trajectory is associated with a success frame $i_{\text{succ}}$, where that frame and all frames after it are labeled as success and the frames before it as failure; for a trajectory that fails entirely or has no success annotation, all frames are labeled $0$,
\begin{equation}
y_i = \mathbb{1}\big[\, i \ge i_{\text{succ}} \,\big].
\label{eq:judge_label}
\end{equation}
Since the success and failure frames within a trajectory are highly imbalanced, we adopt a class-balanced binary cross-entropy as the training objective. Within each batch, we count the numbers of positive and negative frames $N_+$ and $N_-$, and weight the minority class by the inverse of its frequency so that the two classes contribute comparably to the loss,
\begin{equation}
\mathcal{L}_{\text{judge}} = \frac{\sum_i m_i\, w_i\, \mathrm{BCE}\big(\sigma(\hat{s}_i),\, y_i\big)}{\sum_i m_i\, w_i},
\label{eq:judge_loss}
\end{equation}
where $w_i$ is the class-balancing weight described above and $m_i$ is a valid-frame mask.

\textbf{Closed-loop integration with the world model.}\quad
Once trained, the world judge is integrated into the closed loop as a 
reward signal. During a closed-loop rollout, the chunk-wise actions 
output by the policy model are rendered through the unified action map 
and drive the world model to generate chunk by chunk; the video and 
proprioceptive state generated by the world model are returned to the 
policy model for its subsequent prediction, while the world judge 
outputs a success signal for the generated frames, yielding a success 
curve over the rollout. This curve serves both as an automated decision 
for policy evaluation and as dense feedback for reward-driven learning, 
turning policy assessment from reliance on manual inspection into a 
machine-verifiable process.

\begin{figure}[t]
\centering
\includegraphics[width=\linewidth]{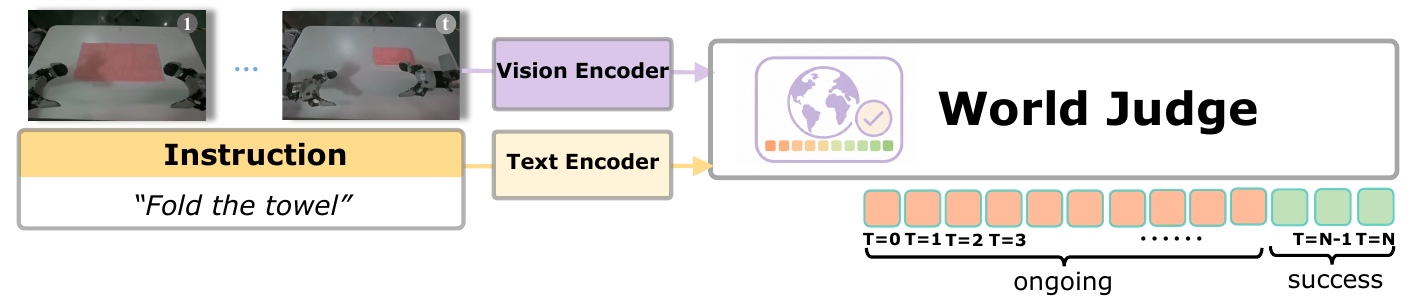} 
\caption{\textbf{World Judge.} The World Judge scores generated rollout frames against task instructions, providing ongoing and success signals for closed-loop policy evaluation. Vision Encoder processes frames, Text Encoder encodes instructions, and the outputs are combined for per-frame success assessment.}
\label{fig:world_judge}
\end{figure}

\subsection{Acceleration}
\label{sec:method_acceleration}

Scalable policy evaluation requires rolling out trajectories chunk by 
chunk and in parallel across many tasks, and the multi-step diffusion 
inference of the world model becomes the throughput bottleneck in this 
setting. We accelerate it from two directions: reducing the number of 
denoising steps required per generation through step distillation, and 
letting a single generation cover a longer time span through 
random-stride training.

\textbf{Step distillation.}\quad
We distill the multi-step diffusion world model into a few-step student 
using the distribution-matching framework of DMD2~\citep{yin2024improved}. 
The distillation involves a frozen teacher, which is the world model 
from the main training stage, a few-step student to be distilled, and a 
trainable fake-score critic that estimates the score of the student's 
output distribution. At each step, the student generates a sample, which 
is re-noised and denoised by the teacher and the critic to yield two 
estimates $x_0^{\text{teacher}}$ and $x_0^{\text{fake}}$; their 
difference forms the distribution-matching gradient that drives the 
student,
\begin{equation}
\nabla_{\!\text{DMD}} = \frac{x_0^{\text{fake}} - 
x_0^{\text{teacher}}}{\big\| x_0^{\text{student}} - x_0^{\text{teacher}} 
\big\|},
\label{eq:dmd_grad}
\end{equation}
where the denominator normalizes the gradient magnitude per sample. The 
student and the critic are optimized alternately so that the critic 
keeps tracking the current student's output distribution. To fit the 
action-conditioned, memory-based setting of GE-Sim 2.0, the teacher, 
student, and critic all receive the same action map condition, and the 
memory frames are kept as clean ground-truth latents throughout; the 
student's number of denoising steps is randomized during training so 
that it supports a flexible one- to few-step configuration at inference. 
We provide the remaining details in the appendix.

\textbf{Temporal acceleration via random-stride training.}\quad
The other direction of acceleration comes from the time span. During 
world-model training, we sample the frames of each chunk with a random 
temporal stride. This exposes the model to trajectories of different 
temporal densities at training time, so that at inference it can perform 
frame skipping, covering up to four times the time span with the same 
number of frames. For the evaluation of long-horizon tasks, this 
significantly reduces the number of autoregressive chunks needed to 
cover a given task segment, with no noticeable loss in spatial-temporal 
consistency.

Combining the two directions, the world model of GE-Sim 2.0 generates a 
100-frame rollout in about 2.3 seconds on a single H100 with only four 
inference steps, reaching a throughput suitable for large-scale parallel 
evaluation.

\begin{figure}[t]
\centering
\includegraphics[width=\linewidth]{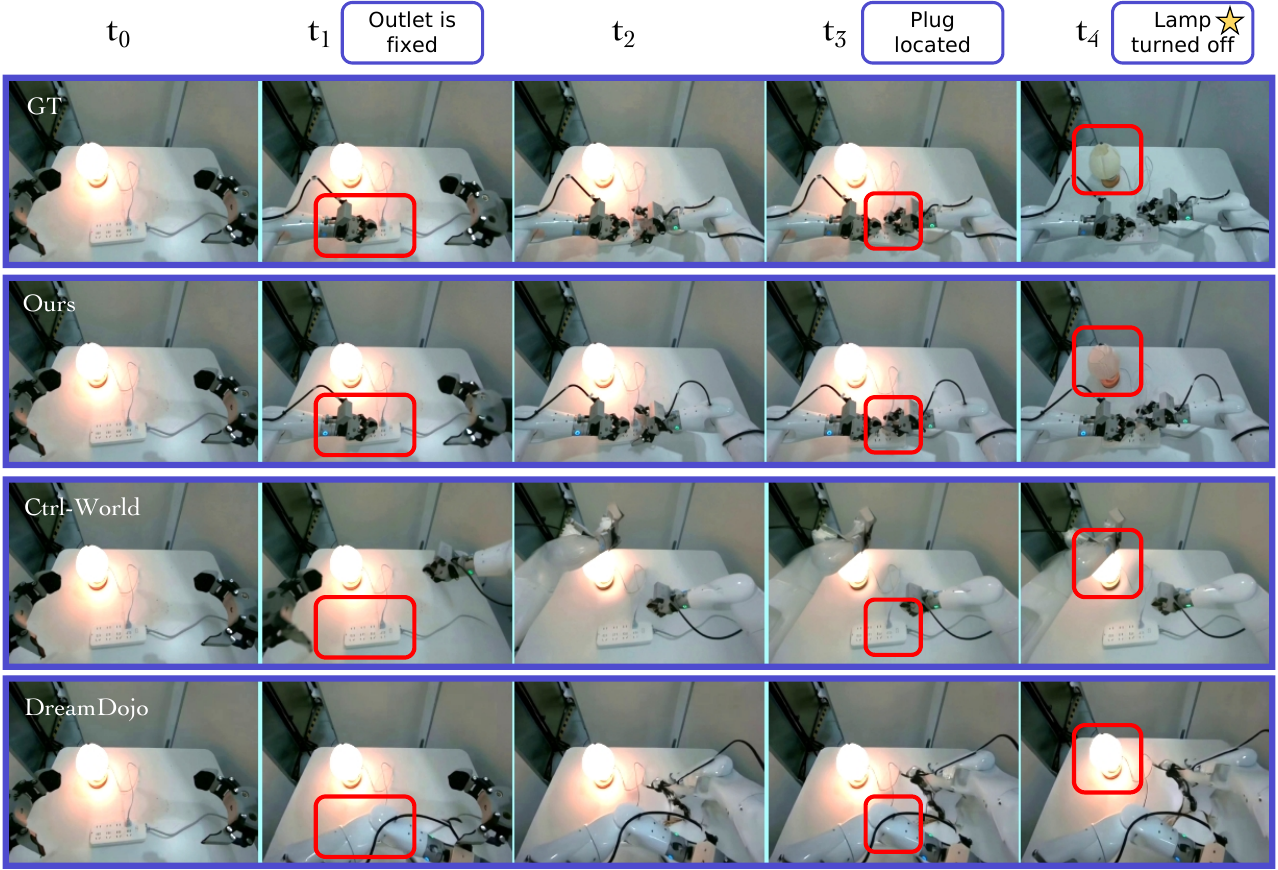}
\caption{\textbf{Qualitative comparison on the \textit{Pull out plug} task.} We compare ground truth (\textit{GT}), GE-Sim 2.0 (\textit{ours}), Ctrl-World, and DreamDojo on an episode where the robot is required to unplug a desk lamp. GE-Sim 2.0 successfully follows the action, removes the plug, and correctly renders the lamp turning off. By contrast, Ctrl-World and DreamDojo both show action-following failures and neither reproduces the lamp-off state after unplugging.}
\label{fig:qualitative_plug}
\end{figure}

\begin{figure}[t]
\centering
\includegraphics[width=\linewidth]{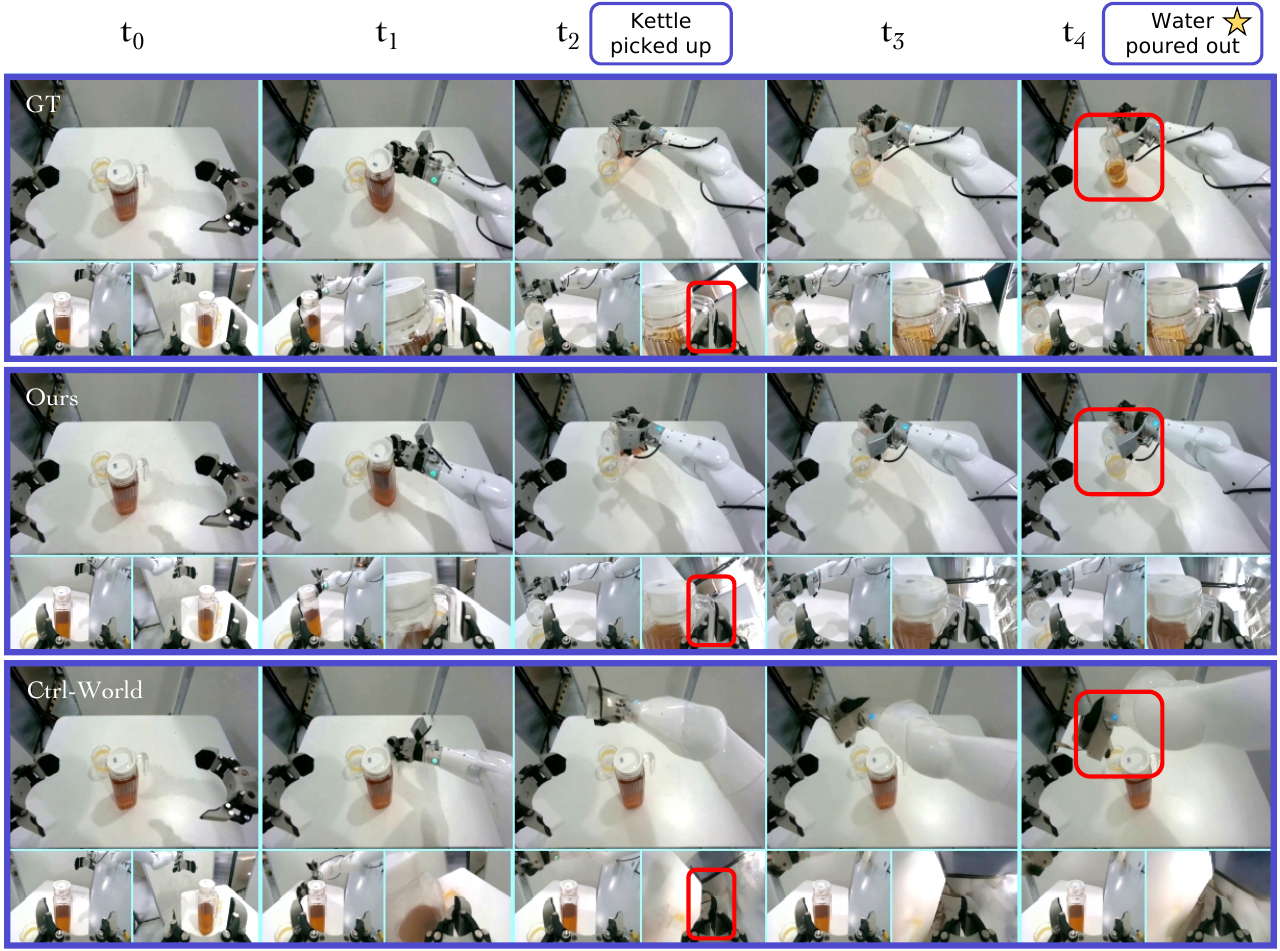}
\caption{\textbf{Qualitative comparison on the \textit{Pour water} task.} We compare ground truth (\textit{GT}), GE-Sim 2.0 (\textit{ours}), and Ctrl-World on an episode where the robot is required to pick up a kettle and pour water into a cup. Each frame contains a top view, with the left view at the lower left and the right view at the lower right. GE-Sim 2.0 successfully follows the action, lifts the kettle, and correctly renders the water-pouring process. By contrast, Ctrl-World shows action-following failures and does not faithfully reproduce the target water-pouring behavior.}
\label{fig:qualitative_pour}
\end{figure}

\begin{table}[t]
\centering
\caption{\textbf{Comparison of robotic world simulators.}
GE-Sim 2.0 supports long-horizon multi-view simulation with proprioceptive
state prediction, pseudo real-time rollout, generalization, and reward
estimation.}
\label{tab:simulator_capability}
\vspace{2pt}
\setlength{\tabcolsep}{5pt}
\renewcommand{\arraystretch}{1.15}
\resizebox{\linewidth}{!}{
\begin{tabular}{lcccccccccc}
\toprule
  
& IRASim
& 1XWM
& GE-Sim 1.0 
& Ctrl-World 
& DreamDojo 
& Interactive WM 
& ABot-PhysWorld
& WorldScape
& MotuBrain
& \textbf{GE-Sim 2.0} \\
\midrule

Long-horizon 
& \textcolor{forestgreen}{\ding{51}}
& \textcolor{red}{\ding{55}} 
& \textcolor{forestgreen}{\ding{51}} 
& \textcolor{red}{\ding{55}} 
& \textcolor{forestgreen}{\ding{51}} 
& \textcolor{forestgreen}{\ding{51}} 
& \textcolor{forestgreen}{\ding{51}}
& \textcolor{forestgreen}{\ding{51}}
& \textcolor{forestgreen}{\ding{51}}
& \textcolor{forestgreen}{\ding{51}} \\

Multi-view 
& \textcolor{red}{\ding{55}}
& \textcolor{red}{\ding{55}}
& \textcolor{forestgreen}{\ding{51}} 
& \textcolor{forestgreen}{\ding{51}} 
& \textcolor{red}{\ding{55}} 
& \textcolor{red}{\ding{55}} 
& \textcolor{red}{\ding{55}}
& \textcolor{red}{\ding{55}}
& \textcolor{forestgreen}{\ding{51}}
& \textcolor{forestgreen}{\ding{51}} \\

Proprioceptive State 
& \textcolor{red}{\ding{55}}
& \textcolor{red}{\ding{55}}
& \textcolor{red}{\ding{55}} 
& \textcolor{red}{\ding{55}} 
& \textcolor{red}{\ding{55}} 
& \textcolor{red}{\ding{55}} 
& \textcolor{red}{\ding{55}}
& \textcolor{red}{\ding{55}}
& \textcolor{red}{\ding{55}}
& \textcolor{forestgreen}{\ding{51}} \\

Pseudo Real-Time 
& \textcolor{red}{\ding{55}}
& \textcolor{red}{\ding{55}}
& \textcolor{red}{\ding{55}} 
& \textcolor{red}{\ding{55}} 
& \textcolor{forestgreen}{\ding{51}} 
& \textcolor{forestgreen}{\ding{51}} 
& \textcolor{red}{\ding{55}}
& \textcolor{forestgreen}{\ding{51}}
& \textcolor{forestgreen}{\ding{51}}
& \textcolor{forestgreen}{\ding{51}} \\

Reward 
& \textcolor{red}{\ding{55}}
& \textcolor{forestgreen}{\ding{51}}
& \textcolor{red}{\ding{55}} 
& \textcolor{red}{\ding{55}} 
& \textcolor{red}{\ding{55}} 
& \textcolor{red}{\ding{55}} 
& \textcolor{red}{\ding{55}}
& \textcolor{red}{\ding{55}}
& \textcolor{red}{\ding{55}}
& \textcolor{forestgreen}{\ding{51}} \\

Open-source
& \textcolor{forestgreen}{\ding{51}}
& \textcolor{red}{\ding{55}}
& \textcolor{forestgreen}{\ding{51}}
& \textcolor{forestgreen}{\ding{51}}
& \textcolor{forestgreen}{\ding{51}}
& \textcolor{forestgreen}{\ding{51}}
& \textcolor{forestgreen}{\ding{51}}
& \textcolor{red}{\ding{55}}
& \textcolor{forestgreen}{\ding{51}}
& \textcolor{forestgreen}{\ding{51}} \\

\bottomrule
\end{tabular}}

\end{table}

\section{Experiments}
\label{sec:exp}

Our experiments are structured to test the central claim of GE-Sim 2.0: that an action-conditioned video model can serve as a \emph{closed-loop} simulator for manipulation policies, not merely a high-quality video generator. Concretely, we ask five questions. \textbf{Q1 (Visual fidelity).} Does GE-Sim 2.0 generate manipulation videos of higher quality than prior robotic world models, both on the broad WorldArena benchmark and on fine-grained per-task replay metrics (\S\ref{sec:exp_video}) \textbf{Q2 (Closed-loop faithfulness).} When a policy model runs against GE-Sim 2.0 rather than the real robot, do the simulated outcomes agree with what happens in the physical world (\S\ref{sec:exp_closedloop})? \textbf{Q3 (Verifiable rewards).} Does the world judge produce success signals that are accurate enough to replace human inspection (\S\ref{sec:exp_reward}). \textbf{Q4 (Policy improvement).} Can the world model and its reward signals be used to generate filtered training data that improves the downstream policy (\S\ref{sec:exp_wmbc}). \textbf{Q5 (Component contribution).} How do the proprioceptive state expert and the specialized reward model affect closed-loop simulation and reward prediction, respectively (\S\ref{sec:exp_ablation})?

\textbf{Tasks.}
We evaluate on six long-horizon dual-arm manipulation tasks chosen to span the failure modes that have historically been hardest for video world models: liquid handling (\textit{Pour water}), deformable-object manipulation (\textit{Fold towels}), fine-force interactions (\textit{Pull out plug}), open-flame ignition (\textit{Borrow flame}), language-grounded selection (\textit{Command grasp \& release}), and surface contact under appearance variation (\textit{Clean mirror stains}).Together, these tasks stress action-following accuracy, contact and deformation rendering, and the ability to faithfully reproduce both policy successes and failures.

\textbf{Baselines.}
For video quality, we compare against Ctrl-World~\citep{guo2025ctrl} and DreamDojo~\citep{gao2026dreamdojo}, two leading open robotic world models that release their model weights and inference code, enabling evaluation under matched action conditions. We additionally place GE-Sim 2.0 on the WorldArena leaderboard~\citep{shang2026worldarena}, which includes recent robotic world models such as GigaWorld~\citep{gigaworld2025}, ABot~\citep{chen2026abot} in both text-driven and action-driven variants, and MotuBrain~\citep{Team2026MotuBrainAA}, as well as closed-source general video generators including Sora~\citep{openai2024sora} and Veo~\citep{deepmind2025veo3}. For the world judge, we compare our specialist reward model against a strong general-purpose multimodal LLM, Qwen3.5-122B~\citep{qwen35blog}, a 122B-parameter mixture-of-experts vision-language model, prompted to classify task success from video.

\textbf{Evaluation protocol.}\quad
For each task we curate 20 evaluation episodes of a $\pi_{0.5}$~\citep{intelligence2025pi05} policy on the real robot. We use two complementary evaluation modes. \textbf{Replay} fixes the action trajectory recorded on the real robot and feeds it into the world model, then measures per-frame and distributional similarity between the simulated and real video; this isolates rendering quality from policy behavior. \textbf{Closed-loop} evaluation reruns the same policy through GE-Sim 2.0 in chunk-wise rollout: the policy generates actions from simulated observations at every step rather than receiving pre-recorded ones, and we report the agreement between simulated and real outcomes; this measures whether the simulator faithfully tracks policy decisions rather than just replaying a fixed sequence. Reward evaluation reports binary classification accuracy (\textit{acc}: whether the model correctly identifies task success) and the frame-index distance between the model's predicted success frame and the human-annotated reference (\textit{dist}), both on world-model rollouts (\textit{WM}) and on ground-truth videos (\textit{GT}).

\subsection{Video Simulation Quality}
\label{sec:exp_video}

\begin{figure}[t]
\centering
\includegraphics[width=\linewidth]{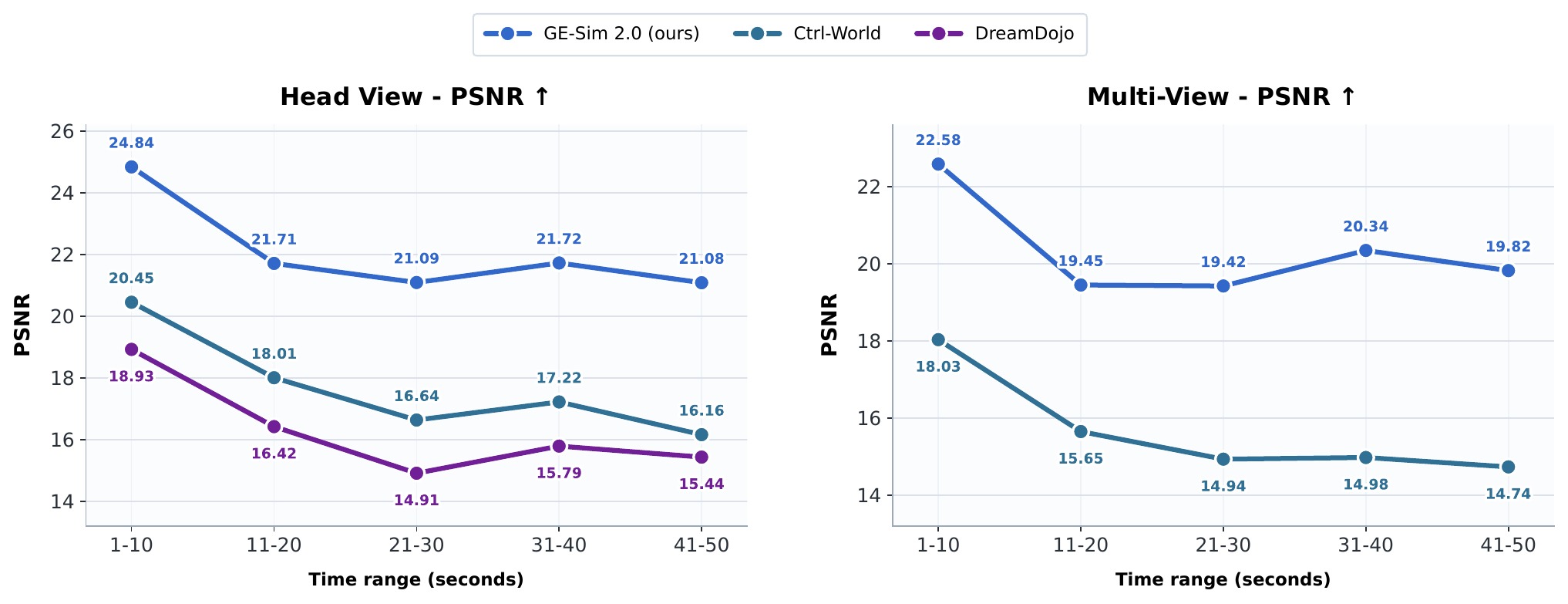}
\caption{Temporal replay quality across different time ranges. We divide each replay video into five consecutive 10-second temporal segments and evaluate PSNR separately for each segment. These results indicate that our method maintain more faithful visual replay over long-horizon rollouts, with a smaller degradation as time progresses.}
\label{fig:grouped_metrics_line_chart}
\end{figure}

\textbf{WorldArena leaderboard.}
Despite using only a 2B-parameter backbone, GE-Sim 2.0 attains the top overall score on the public WorldArena leaderboard, outperforming both dedicated robotic world models (Ctrl-World, DreamDojo, GigaWorld, ABot in its text and action variants) and substantially larger closed-source general video generators including Sora and Veo. This suggests that domain-specific training on diverse, action-grounded robot data is more important for manipulation simulation than raw model scale alone.

\textbf{Per-task replay metrics.}
Table~\ref{tab:replay} reports replay quality averaged over the six tasks, separately for the head camera and for the joint multi-view configuration. GE-Sim 2.0 dominates across all five metrics in both views. On the head view, it improves PSNR by $+3.96$\,dB over Ctrl-World and $+5.67$\,dB over DreamDojo, halves FID (32.3 vs.\ 62.7), and reduces FVD by more than a factor of two (481 vs.\ 1084). On the considerably harder multi-view setting, where temporal coherence must hold simultaneously across head, left-wrist, and right-wrist cameras, the improvement is even more pronounced: $-31.9$\,FID and $-2.5\times$\,FVD relative to Ctrl-World, while DreamDojo did not produce comparable multi-view outputs. The same ranking holds task by task: GE-Sim 2.0 is the best simulator on every individual task across all five metrics. The consistent improvement across benchmarks and all task subsets indicates that the quality gains are broad-based rather than artefact-specific.

\textbf{Long-horizon temporal robustness.}
We further compare GE-Sim 2.0 against Ctrl-World and DreamDojo on visual stability over long-horizon rollouts. The aggregate replay metrics above average over the full 50-second rollout and can mask whether visual quality remains stable as the number of autoregressive chunks grows. To probe this, we divide each replay video into five consecutive 10-second segments and report PSNR per segment (Figure~\ref{fig:grouped_metrics_line_chart}). The two views tell the same story. On the head view, GE-Sim 2.0 decays from $24.84$ to $21.08$\,dB, a swing of under $4$\,dB across the full rollout, while Ctrl-World and DreamDojo drop sharply after the first segment and end at $16.16$ and $15.44$\,dB, nearly $5$\,dB below ours. On the harder multi-view setting, GE-Sim 2.0 stays above $19.4$\,dB throughout, whereas Ctrl-World falls to the $15$\,dB range by the second segment and continues to decline. The shape of the curves matters as much as the absolute gap: GE-Sim 2.0 flattens after the first segment while the baselines keep degrading, so the advantage widens as rollouts get longer. This long-horizon stability is precisely what a simulator needs for closed-loop evaluation, and we attribute it to the large-scale real-robot retraining, the memory-frame augmentation, and the random-stride training of \S\ref{sec:method_acceleration}.

\begin{table}[t]
\centering
\caption{\textbf{Replay visual quality} averaged over six manipulation tasks. GE-Sim 2.0 outperforms both robotic world model baselines on all five metrics, in both single-view and multi-view settings. Best in \textbf{bold}; ``—'' denotes no comparable multi-view output.}
\label{tab:replay}
\vspace{2pt}
\setlength{\tabcolsep}{4pt}
\resizebox{\linewidth}{!}{
\begin{tabular}{lrrrrr rrrrr}
\toprule
& \multicolumn{5}{c}{Head View} & \multicolumn{5}{c}{Multi-View} \\
\cmidrule(lr){2-6}\cmidrule(lr){7-11}
Method & PSNR$\uparrow$ & SSIM$\uparrow$ & LPIPS$\downarrow$ & FID$\downarrow$ & FVD$\downarrow$ & PSNR$\uparrow$ & SSIM$\uparrow$ & LPIPS$\downarrow$ & FID$\downarrow$ & FVD$\downarrow$ \\
\midrule
Ctrl-World~\citep{guo2025ctrl}     & 19.09 & 0.786 & 0.296 & \phantom{0}62.70 & 1083.7 & 16.65 & 0.735 & 0.407 & \phantom{0}56.85 & 1527.5 \\
DreamDojo~\citep{gao2026dreamdojo} & 17.38 & 0.728 & 0.353 & \phantom{0}79.82 & 1155.0 & — & — & — & — & — \\
GE-Sim 2.0 (ours)                  & \textbf{23.05} & \textbf{0.846} & \textbf{0.145} & \phantom{0}\textbf{32.28} & \phantom{0}\textbf{481.3} & \textbf{20.80} & \textbf{0.796} & \textbf{0.217} & \phantom{0}\textbf{24.92} & \phantom{0}\textbf{613.5} \\
\bottomrule
\end{tabular}}
\end{table}

\begin{figure}[!htbp]
\centering
\includegraphics[width=\linewidth]{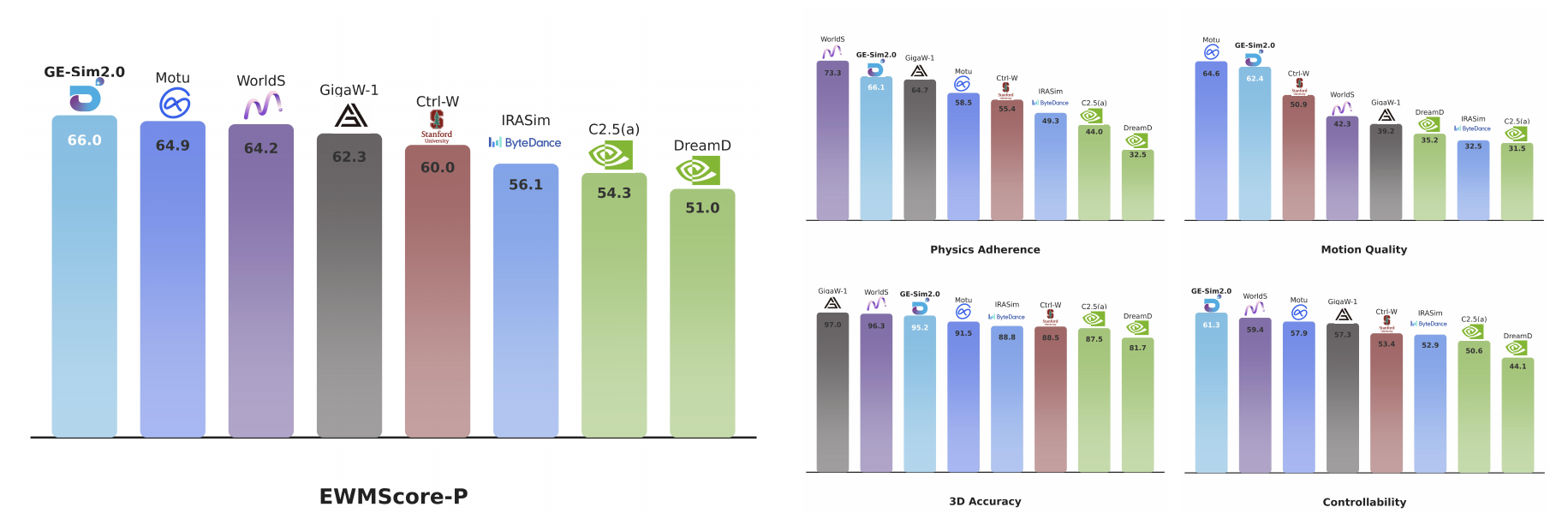}
\caption{\textbf{WorldArena leaderboard.} 
GE-Sim 2.0 achieves the top overall score on the WorldArena benchmark, 
outperforming prior robotic world models and closed-source general video 
generators.}
\label{fig:worldarena}
\end{figure}

\subsection{Closed-Loop Policy Consistency}
\label{sec:exp_closedloop}

Replay fidelity does not by itself imply that a world model can serve as a policy simulator. In closed-loop use, small errors in visual state, proprioceptive state, or contact evolution can change the policy's subsequent actions and lead to different task outcomes. We therefore run the same $\pi_{0.5}$ policy inside each world model and compare the resulting simulated outcomes with physical-robot outcomes under matched task settings.

\begin{figure}[t]
    \centering
    \includegraphics[width=1.0\linewidth]{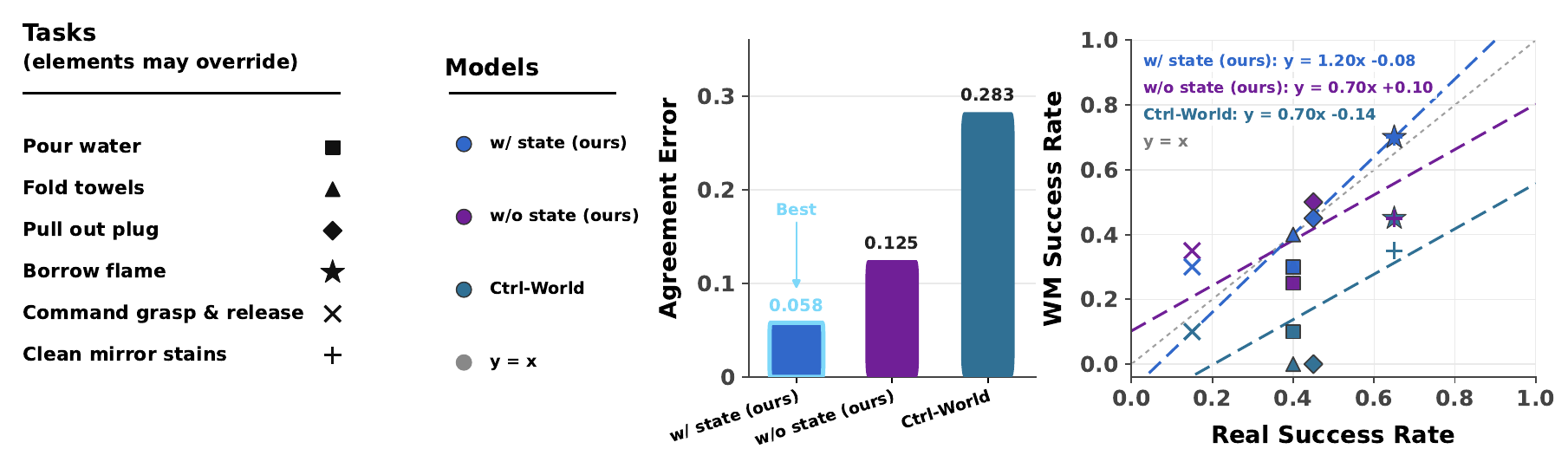}
    \caption{
    \textbf{World-model success-rate alignment with real-robot outcomes.}
    We compare the task success rates measured in the real world and those predicted by different world models under closed-loop policy rollouts.
    Each marker denotes one manipulation task, while different dashed lines correspond to Ctrl-World, our model without state conditioning, and our full model with state conditioning.
    The gray dashed line indicates perfect agreement ($y=x$) between world-model and real-world success rates.
    Compared with Ctrl-World, our models show stronger alignment with real-world policy performance, and state conditioning further improves the consistency between simulated and physical outcomes.
    }
    \label{fig:wm_real_success_alignment}
\end{figure}

\textbf{Task-level success-rate alignment.}
Figure~\ref{fig:wm_real_success_alignment} compares the real-robot success rate with the success rate measured inside each world model across the six tasks. Each marker denotes one task, and the diagonal indicates perfect agreement. GE-Sim 2.0 with state conditioning shows the closest task-level alignment: its fitted trend has a slope close to one with a small negative offset, suggesting that it better preserves the relative task difficulty observed on the real robot. In contrast, removing state conditioning weakens this correspondence, yielding a shallower trend and a positive offset. Ctrl-World shows the largest mismatch, with a similarly shallow slope but a negative offset, indicating a tendency to underestimate policy success across tasks.

This result shows that state-aware simulation improves aggregate closed-loop calibration. The remaining gaps are most visible on contact-sensitive tasks such as \textit{Pull out plug} and \textit{Command grasp \& release}, where small errors in grasp state, contact timing, or object interaction can alter the downstream policy rollout. Thus, task-level success alignment provides a coarse but important measure of whether the simulator preserves real-world policy performance trends.

\textbf{Episode-level outcome agreement.}
Task-level success rates can still hide case-level mismatches: two simulators may have similar marginal success rates while predicting success on different initial conditions. We therefore additionally evaluate closed-loop agreement as an episode-level binary prediction problem, where a simulated success or failure is compared with the corresponding physical-robot outcome.

Figure~\ref{fig:confusion_matrix} reports the per-task confusion matrices for the three world models, and summarizes accuracy, precision, and recall. Accuracy measures the fraction of matched simulated and real outcomes, precision measures how often simulated successes are real successes, and recall measures how many real successes are recovered by the simulator. GE-Sim 2.0 with state conditioning obtains the best average accuracy and recall: accuracy improves from $0.63$ for Ctrl-World and $0.74$ without state conditioning to $0.81$, while recall improves from $0.25$ and $0.67$ to $0.82$. The large recall gain indicates that the state-conditioned simulator is substantially better at preserving episodes where the real policy succeeds, rather than only matching the marginal success frequency.

The task-wise confusion matrices reveal both strengths and limitations. GE-Sim 2.0 recovers more true-positive outcomes on tasks such as \textit{Fold towels}, \textit{Borrow flame}, and \textit{Clean mirror stains}, but false positives and false negatives remain on contact-rich tasks. These errors suggest that fine-grained contact state and long-horizon state accumulation are still challenging. Overall, the task-level and episode-level results together support GE-Sim 2.0 as a more faithful closed-loop evaluation environment than the compared baselines.

\subsection{World Judge: Reward Quality}
\label{sec:exp_reward}

Beyond rendering rollouts, GE-Sim 2.0 must also \emph{score} them: closed-loop policy evaluation and reward-driven learning require machine-verifiable success signals. We evaluate our world judge against a strong general VLM baseline, Qwen3.5-122B, on two complementary inputs: simulated rollouts (\textit{WM}), which is the operating mode used at policy-evaluation time, and ground-truth videos (\textit{GT}), which probes upper-bound reward accuracy without simulator-induced artifacts. Both models receive the same per-frame multi-image input for a fair comparison.

Table~\ref{tab:reward} summarizes the results. On WM rollouts, our world judge reaches $79\%$ accuracy versus $60\%$ for Qwen ($+19$\,pp), and reduces event distance from $57.8$ to $28.2$ frames. The advantage carries over to GT videos: $87\%$ versus $58\%$ ($+29$\,pp), with event distance reduced from $64.7$ to $15.7$ frames. The accuracy ranking holds on five of the six tasks; the only exception is \textit{Command grasp \& release}, a task dominated by discrete object-grasping events that a general VLM is comparatively well suited to recognize. Qwen further fails outright to estimate success frames on \textit{Clean mirror stains}, a task whose success criterion involves subtle appearance changes rather than discrete object events. The remaining gaps on \textit{Pour water} and \textit{Borrow flame} point to the difficulty of detecting small-scale visual signals such as liquid level and flame transfer, which remains an open challenge for task-aware reward models. The gap between WM-mode and GT-mode accuracy ($\sim8$\,pp) reflects the share of error attributable to simulator artifacts rather than inherent task ambiguity, and is small enough that reward signals from GE-Sim 2.0 remain reliable for policy assessment.

\begin{table}[t]
\centering
\caption{\textbf{World judge quality}: success-classification accuracy (\textit{acc}) and event distance in frames (\textit{dist}) on world-model rollouts (\textit{WM}) and ground-truth videos (\textit{GT}). Our world judge reports consensus predictions; Qwen produces a single per-video prediction. ``—'' denotes that the model failed to identify a success frame. Our world judge outperforms the general VLM baseline by a large margin on most tasks.}
\label{tab:reward}
\vspace{2pt}
\setlength{\tabcolsep}{4pt}
\begin{tabular}{lcccccccc}
\toprule
& \multicolumn{4}{c}{Ours} & \multicolumn{4}{c}{Qwen3.5-122B-A10B} \\
\cmidrule(lr){2-5}\cmidrule(lr){6-9}
& \multicolumn{2}{c}{WM} & \multicolumn{2}{c}{GT} & \multicolumn{2}{c}{WM} & \multicolumn{2}{c}{GT} \\
Task & acc$\uparrow$ & dist$\downarrow$ & acc$\uparrow$ & dist$\downarrow$ & acc$\uparrow$ & dist$\downarrow$ & acc$\uparrow$ & dist$\downarrow$ \\
\midrule
Borrow flame              & \textbf{.96} & \textbf{37.0} & \textbf{.95} & \textbf{24.4} & .71 & \phantom{0}96.5 & .70 & 115.3 \\
Clean mirror stains       & \textbf{.82} & \textbf{\phantom{0}9.3} & \textbf{.90} & \textbf{12.4} & .50 & — & .50 & — \\
Command grasp \& release  & .72 & \textbf{\phantom{0}7.8} & .80 & \textbf{\phantom{0}4.0} & \textbf{.81} & \phantom{0}38.7 & \textbf{.85} & 43.4 \\
Fold towels               & \textbf{.65} & \textbf{13.0} & \textbf{.60} & \textbf{\phantom{0}9.5} & .63 & \phantom{0}38.0 & .50 & 30.5 \\
Pour water                & \textbf{.90} & 61.4 & \textbf{1.00} & 26.2 & .55 & \textbf{\phantom{0}22.0} & .35 & \textbf{13.0} \\
Pull out plug             & \textbf{.80} & \textbf{24.2} & \textbf{.95} & \textbf{\phantom{0}7.5} & .55 & \phantom{0}34.0 & .60 & 17.0 \\
\midrule
\textbf{Average}          & \textbf{.79} & \textbf{28.2} & \textbf{.87} & \textbf{15.7} & .60 & 57.8 & .58 & 64.7 \\
\bottomrule
\end{tabular}
\end{table}

\begin{figure}[t]
    \centering
    \includegraphics[width=1.0\linewidth]{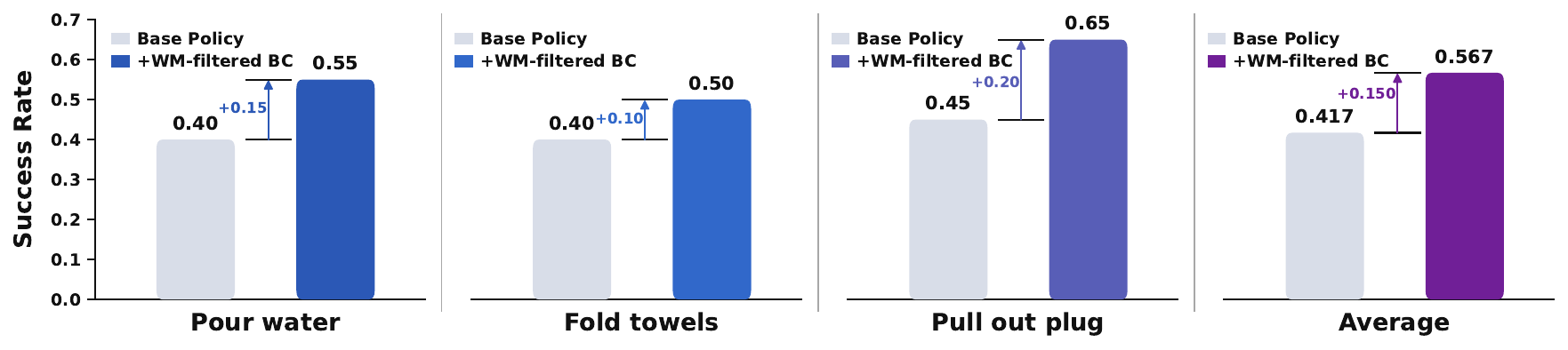}
    \caption{
    \textbf{Policy improvement with WM-filtered behavior cloning.}
success rates of the $\pi_{0.5}$ policy before and after augmenting the original training data with filtered synthetic trajectories generated by GE-Sim 2.0.
    For each task, we run the policy inside the world model, score the generated rollouts with our reward model, retain high-reward trajectories, and mix them with the original behavior cloning data for policy training. 
    WM-filtered BC consistently improves policy success rates across all evaluated contact-rich manipulation tasks.
    }
    \label{fig:wm_filtered_bc}
\end{figure}

\subsection{WM-Based Filtered Behavior Cloning}
\label{sec:exp_wmbc}

Beyond evaluation, GE-Sim 2.0 can also be used as a data-generation engine for policy improvement. As a proof of concept, we run $\pi_{0.5}$ inside GE-Sim 2.0, score the generated rollouts with the world judge, retain trajectories whose reward exceeds a per-task threshold, and mix the retained trajectories with the original behavior-cloning data. Due to the cost of physical-robot evaluation, we evaluate this procedure on three representative tasks covering liquid handling, deformable-object manipulation, and fine-force interaction.

Figure~\ref{fig:wm_filtered_bc} reports real-robot success rates before and after adding the filtered world-model data. WM-filtered BC improves success from $0.40$ to $0.55$ on \textit{Pour water}, from $0.40$ to $0.50$ on \textit{Fold towels}, and from $0.45$ to $0.65$ on \textit{Pull out plug}. The average success rate increases from $0.417$ to $0.567$, corresponding to an absolute gain of $0.150$. These results indicate that GE-Sim 2.0 can produce synthetic rollouts that are useful not only for evaluation, but also for improving downstream policy learning after reward-based filtering.

\begin{figure}[t]
    \centering
    \includegraphics[width=\linewidth]{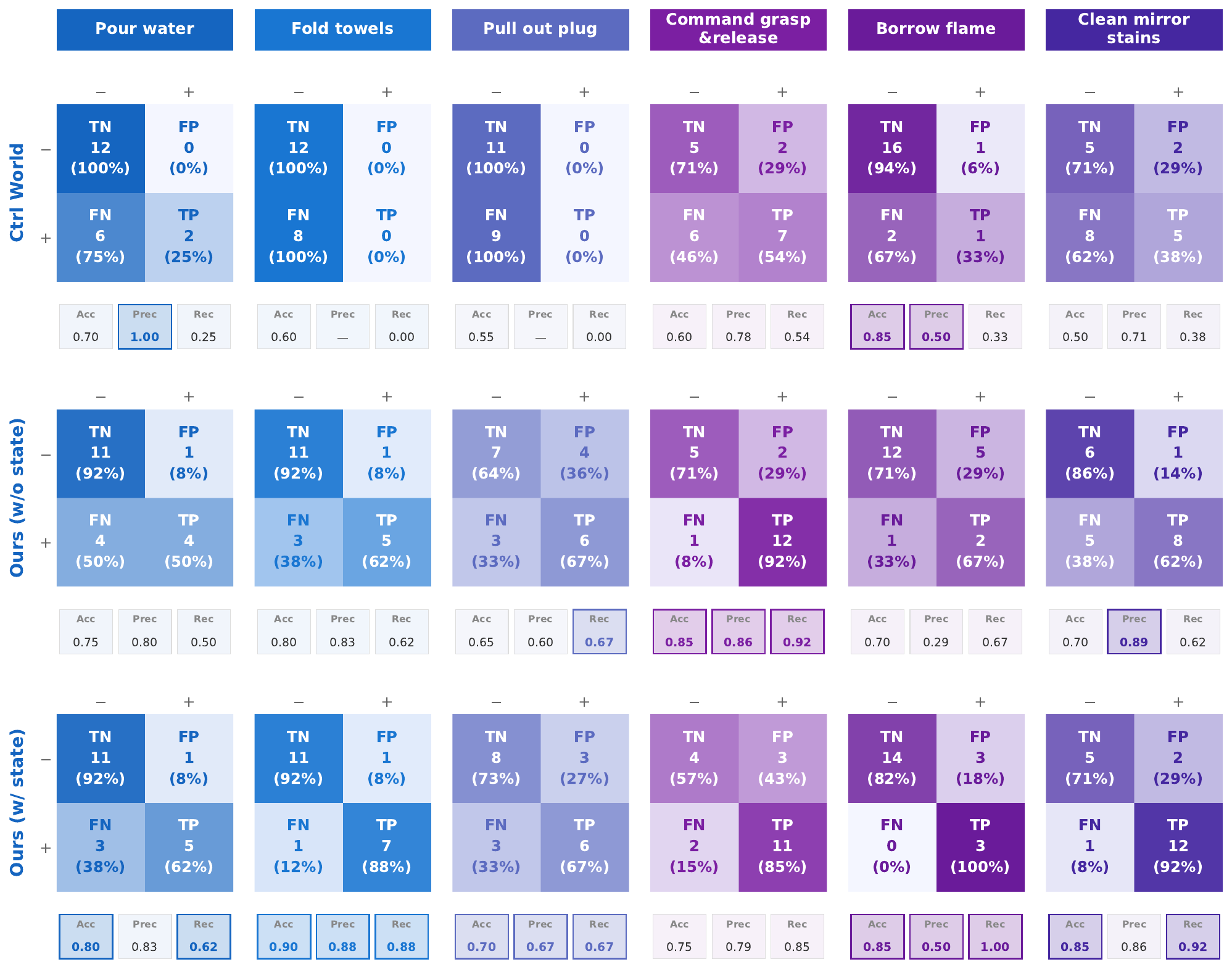}
    \caption{Confusion matrices evaluating the agreement between closed-loop world-model simulations and physical-robot outcomes for binary task success/failure. Compared with Ctrl-World, our model achieves higher true-positive rates, suggesting better preservation of real-world task success patterns, especially in contact-rich scenarios.}
    \label{fig:confusion_matrix}
\end{figure}

\subsection{Ablation Studies}
\label{sec:exp_ablation}

We isolate the contribution of two key components of GE-Sim 2.0. The proprioceptive state expert is evaluated through closed-loop sim-to-real agreement, while the specialist world judge is evaluated through reward-prediction accuracy and event localization.

\textbf{Proprioceptive state expert.} 
We evaluate the impact of the proprioceptive state expert by comparing closed-loop world-model rollouts with and without it against real robot outcomes, using binary task success/failure. As shown in Figure~\ref{fig:confusion_matrix}, removing the state expert reduces the average episode-level accuracy from $0.81$ to $0.74$ and recall from $0.82$ to $0.67$, while precision remains largely unchanged. The confusion matrices highlight that including the state expert substantially improves the model's ability to correctly recover true-positive outcomes, particularly for contact-rich tasks such as \textit{Fold towels} and \textit{Pull out plug}.

The gains are most pronounced on tasks that require accurate state tracking across multiple action chunks, such as \textit{Fold towels}, \textit{Borrow flame}, and \textit{Clean mirror stains}. These tasks involve fine-grained arm motion, contact maintenance, or long-horizon accumulation, where the robot's actual joint state can deviate from the commanded action due to compliance and contact forces. Feeding back the decoded proprioceptive state helps prevent this drift from compounding during autoregressive rollout.

A small reversal appears on \textit{Command grasp \& release}, where the model without state conditioning obtains slightly higher accuracy, suggesting that this task is less dependent on precise joint-state recovery and more dominated by coarse visual progress. Overall, the ablation confirms that the proprioceptive state expert improves closed-loop consistency by providing policy-relevant state feedback beyond action-conditioned video prediction alone.

\textbf{Specialist vs.\ general-purpose reward model.}
Replacing our world judge with the Qwen3.5-122B VLM (Table~\ref{tab:reward}) reduces WM-mode accuracy by $19$\,pp and increases event distance by $30$ frames, with similar gaps on GT videos. Qwen further fails to identify a success frame on \textit{Clean mirror stains}, indicating that prompting a general VLM is not a robust substitute for a task-aware specialist judge. Together, these results justify the small specialist reward head as a core component of the closed-loop simulator.
\section{Conclusion}
\label{sec:conclusion}

We presented GE-Sim 2.0, a step from a ``view-only'' video world simulator toward an omni world simulator for robotic manipulation. On top of the action-conditioned video generation framework of Genie Envisioner, retrained on thousands of hours of real-robot data, three coordinated components close the loop: a proprioceptive state expert that decodes joint-space state from the video latents, a world judge that turns rollouts into machine-verifiable success signals, and a DMD-based acceleration framework that compresses multi-step diffusion into few-step inference. Together, they integrate visual simulation, proprioceptive state prediction, and reward assessment into a single efficient framework. On six long-horizon dual-arm tasks, GE-Sim 2.0 leads on both the WorldArena leaderboard and multi-view replay metrics, its closed-loop success rates agree with the real robot within 1\,pp, the world judge clearly outperforms a strong general VLM, and filtered behavior cloning on its rollouts and rewards yields a 15\,pp average gain on the real-robot policy. These results advance the video world simulator from a forward predictor into a closed-loop platform with built-in evaluation, offering a usable starting point for scalable evaluation and closed-loop learning of manipulation policies.

\section{Looking Forward}
\label{sec:looking_forward}

GE-Sim 2.0 explores one viable instantiation of an omni world simulator, defining the components such a system should include and demonstrating competitive simulation on robotic manipulation. Moving toward a world simulator for general embodied AI leaves several directions open.

\textbf{Data and model scaling.}\quad Our current focus is on methodology, and the foundation model is trained on dual-arm teleoperation and on-robot replay from a fixed embodiment. A general embodied world simulator will require extending to large-scale cross-embodiment, ego-view, and UMI-style data alongside broader manipulation video, with a systematic study of data- and model-scaling behavior. Pushing the foundation backbone is a core bottleneck on the path to general embodied simulation.

\textbf{From an external judge to a unified model.}\quad The current world judge is an external module that suffices for task-success assessment. A longer-term form is to unify state decoding and success judgment into a single world model whose forward pass jointly produces future frames, proprioceptive state, and task-completion signals, removing the separate reward network and letting reward signals inherit the priors learned by the world model at 
scale.

\textbf{From offline filtered BC to online closed-loop learning.}\quad Our experiments show that the rollouts and rewards of GE-Sim 2.0 can already improve a real-robot policy through filtered behavior cloning. A natural continuation is online policy learning and reinforcement learning inside the simulator, advancing the system from an evaluator and data filter into an environment for policy training. The closed-loop and ablation studies here are conducted with a single VLA policy family; validating GE-Sim 2.0 across a broader range of policies is a useful 
next step.

\section{Acknowledgements}
\label{sec:acknowledgements}

\noindent
We thank Aogeliqiang Niyazi, Wei Xu, Wenhao Wang, Ruofei Niu,
Jingyuan Wang, Xiongfeng Cai, Haoyu Cao, Cheng Jing, Pengfei Zhou, Donglin Yang,
Liyan Zhang, Xiaoyun Hu, Feng Han, Chuankang Li, Sukai Wang,
Yuxiang Yan, Jia Zeng, Yuehan Niu, Xuan Hu, and Jing Wu
for their valuable support and contributions to this project.

{
\small
\bibliography{ref}
}

\clearpage
\section{Appendix}

\subsection{Memory-Frame Augmentation Details}
\label{sec:appendix_mem_aug}

This section provides the augmentation settings referenced in Section~\ref{sec:method_foundation}. At inference, memory frames come from the model's own rollouts and drift away from the clean memory frames seen during training. We apply three types of perturbations to memory-frame latents at training time, all gated by an outer activation probability of $0.8$. \emph{Progressive noise mixing} mixes Gaussian noise into the memory latents along the time axis, with a per-frame activation probability of $0.5$ and a per-frame perturbation scale of $\sigma_{\text{mem}} = 0.2$; a separate first-frame perturbation is activated with probability $0.2$ at scale $\sigma_{\text{first}} = 0.5$. \emph{Local Gaussian blur} is applied with probability $0.5$, using a kernel size drawn from $[1, 5]$ and $\sigma$ drawn from $[0.1, 1.3]$, restricted to a connected-component mask covering roughly $20\%$ of the frame. \emph{Multi-view-synchronized color jitter} is applied with probability $0.3$, sharing the same jitter across all camera views to keep lighting and color statistics consistent across head and wrist views.

\subsection{Proprioceptive State Expert Details}
\label{sec:appendix_state_expert}

\textbf{History-state augmentation.}\quad
The two perturbations referenced in Section~\ref{sec:method_state_expert} are applied independently, each with probability $0.5$. The \emph{delta-index shift} shifts the entire history delta index by an integer $\Delta \sim \mathrm{Uniform}\{-3, \ldots, -1, 1, \ldots, 3\}$ and clips it to the valid range, simulating a one- to three-frame temporal misalignment between the policy and the simulator. The \emph{temporal resampling} downsamples the history segment from $n_{\text{prev}}$ to $n_{\text{prev}} - 1$ frames and linearly interpolates it back to $n_{\text{prev}}$, acting as a low-pass distortion that preserves the long-term trend while removing single-frame detail. Both perturbations are disabled at validation time.

\subsection{DMD2 Acceleration Details}
\label{sec:appendix_dmd2}

This section provides the implementation details referenced in Section~\ref{sec:method_acceleration}. The teacher is the vision expert from the main training stage and is frozen throughout. The student and the fake-score critic are both initialized from the teacher. The student targets 4 inference steps, with the actual number of denoising steps randomized between 1 and 4 during training so that one- to few-step inference is supported.

We update the student once every 5 steps and the critic on the remaining four, with a single-step warmup before the first student update. The critic loss is weighted by $1/\sigma^2$. The student is trained with the fixed sigma schedule $[1.0, 0.9375, 0.8333, 0.625]$, concentrated near full noise. During generator updates we shift the sigma distribution toward medium-to-high noise (shift = 3) to avoid wasted near-zero-noise batches; during critic updates we shift it further toward $\sigma \approx 1$ (shift = 5). The student is optimized with AdamW ($\beta_2 = 0.999$), weight decay $0.01$, and 100 warmup steps; the critic uses a smaller learning rate of $2 \times 10^{-6}$ with $\beta_1 = 0$, $\beta_2 = 0.999$, and weight decay $0.01$. No auxiliary flow-matching loss is added to the student.

\subsection{Additional Qualitative Results}
\label{sec:appendix_qualitative}

We provide additional qualitative comparisons across all six manipulation tasks. Figures~\ref{fig:more_results_grasp_three}, \ref{fig:more_results_mirror_three}, \ref{fig:more_results_candle_three}, \ref{fig:more_results_towel_three}, and~\ref{fig:more_results_plug_three} show multi-view rollouts (head together with the left and right wrist cameras) comparing GE-Sim 2.0 against Ctrl-World on Command grasp \& release, Clean mirror stains, Borrow flame, Fold towels, and Pull out plug. Figures~\ref{fig:more_results_water_head}, \ref{fig:more_results_grasp_head}, \ref{fig:more_results_candle_head}, and~\ref{fig:more_results_mirror_head} show head-view comparisons on Pour water, Command grasp \& release, Borrow flame, and Clean mirror, the four tasks where DreamDojo also produces a comparable single-view output.

\label{sec:appendix}

\begin{center}
    \centering
 \vspace{-3mm}
    \includegraphics[width=0.88\linewidth]{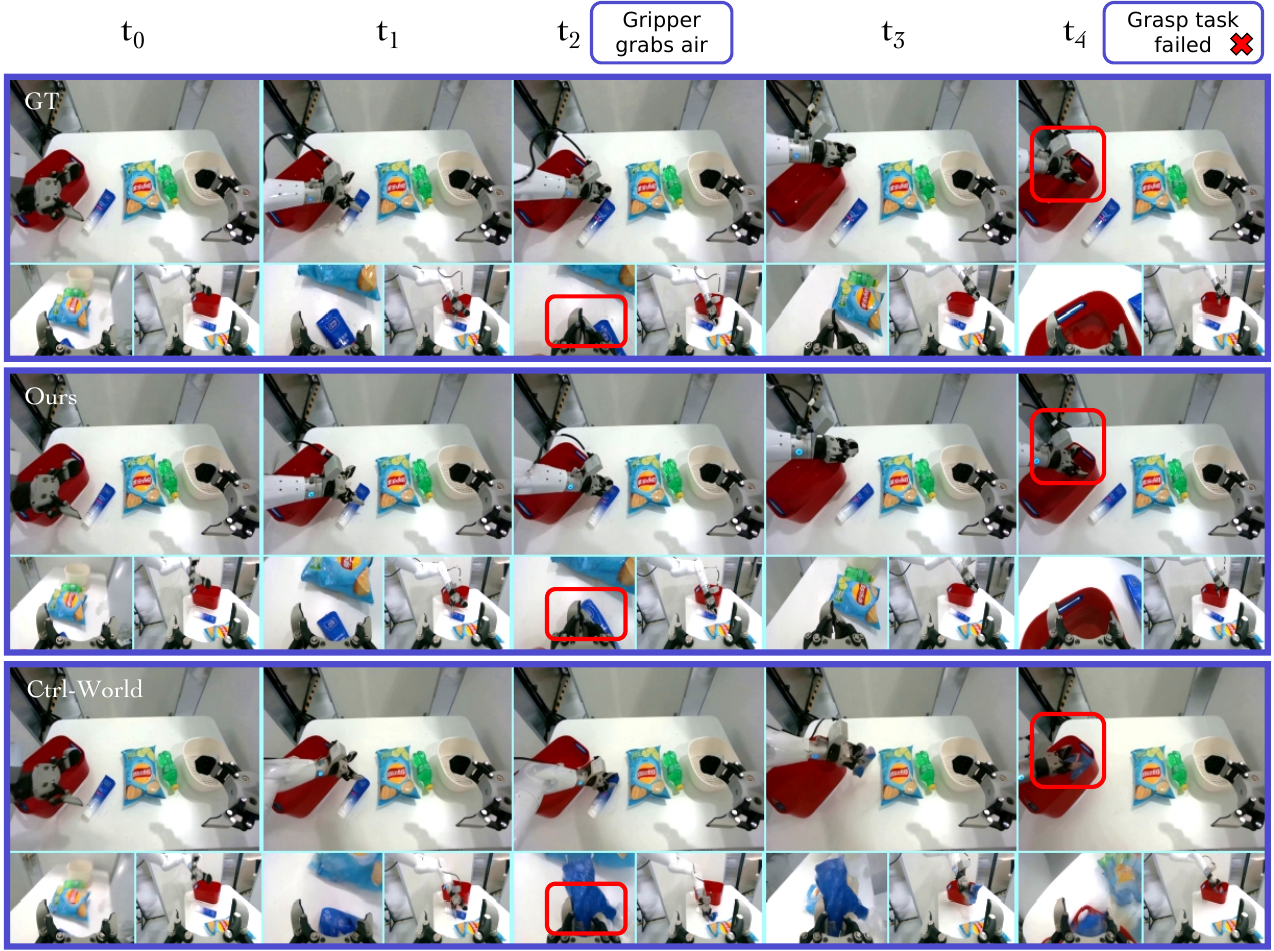}
\vspace{-1mm}
    \captionof{figure}{Command grasp \& release (Multi-View).}
    
    \label{fig:more_results_grasp_three}
\end{center}

\begin{center}
    \centering
\vspace{-2mm}
    \includegraphics[width=0.88\linewidth]{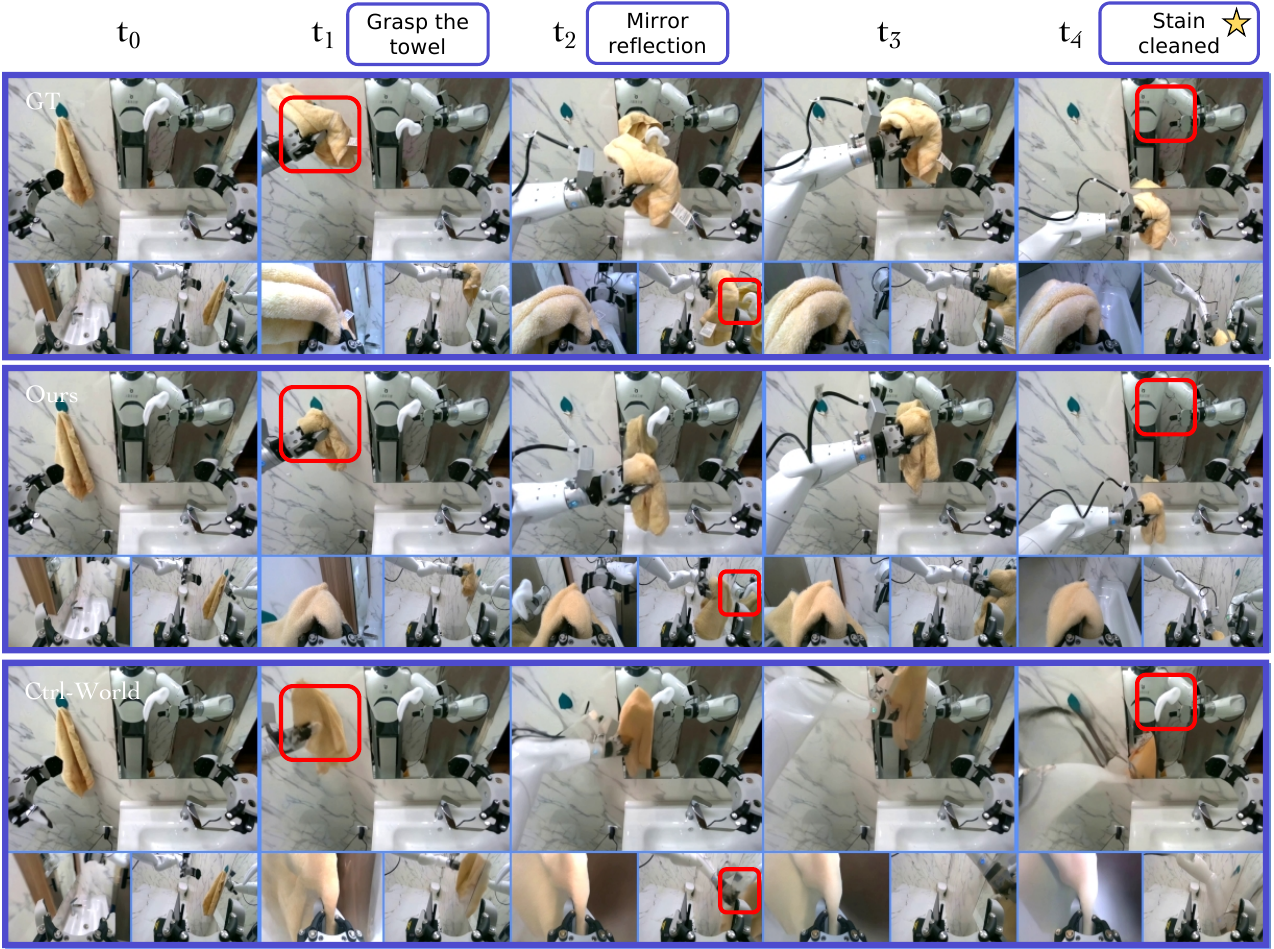}
    \vspace{-1mm}
    \captionof{figure}{Clean mirror stains (Multi-View).}
    \label{fig:more_results_mirror_three}
\end{center}

\begin{center}
    \centering
    \includegraphics[width=0.88\linewidth]{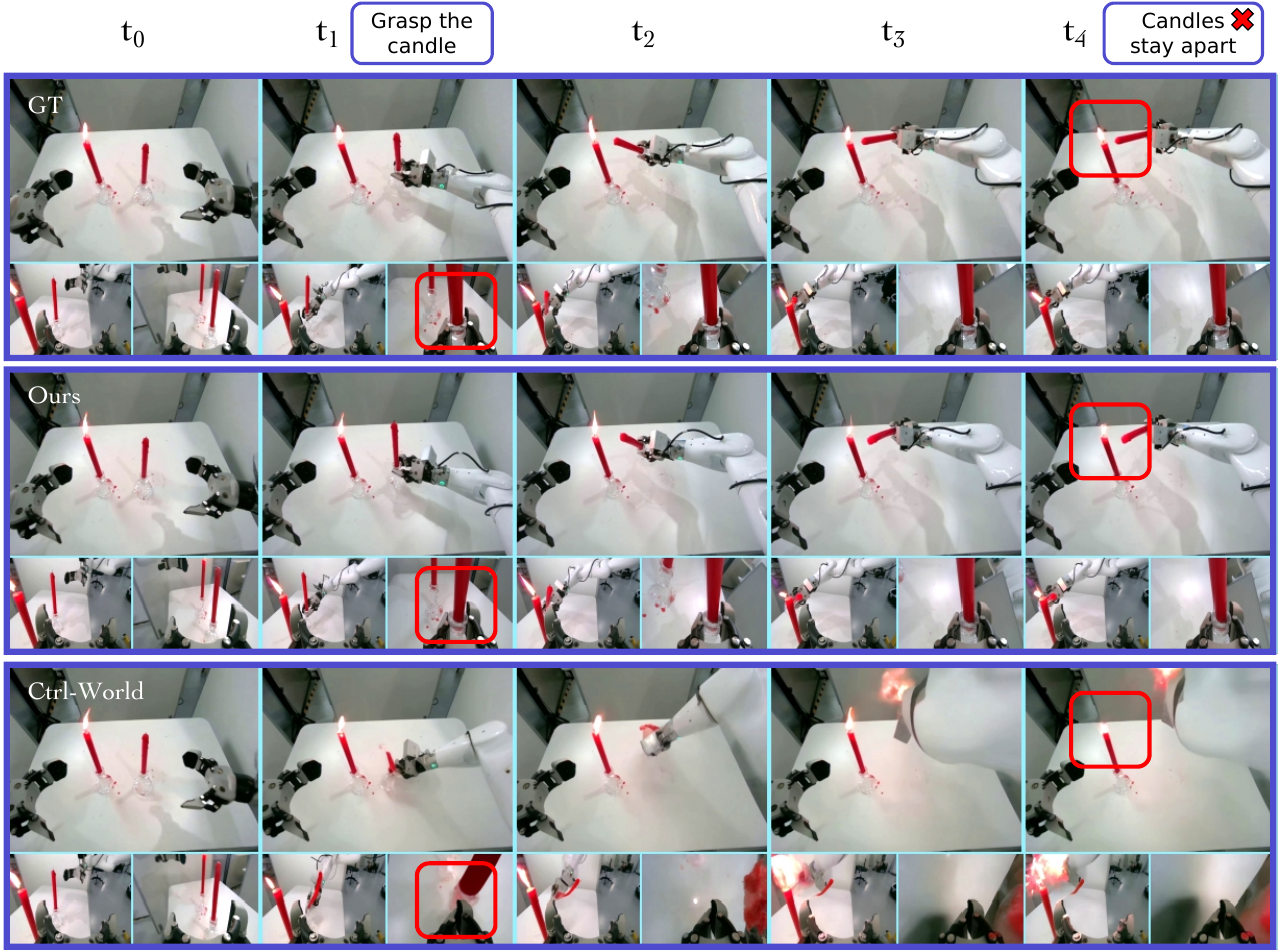}
    \vspace{-2mm}
    \captionof{figure}{Borrow flame (Multi-View).}
    \label{fig:more_results_candle_three}
\end{center}

\begin{center}
    \centering
    \vspace{-1mm}
    \includegraphics[width=0.88\linewidth]{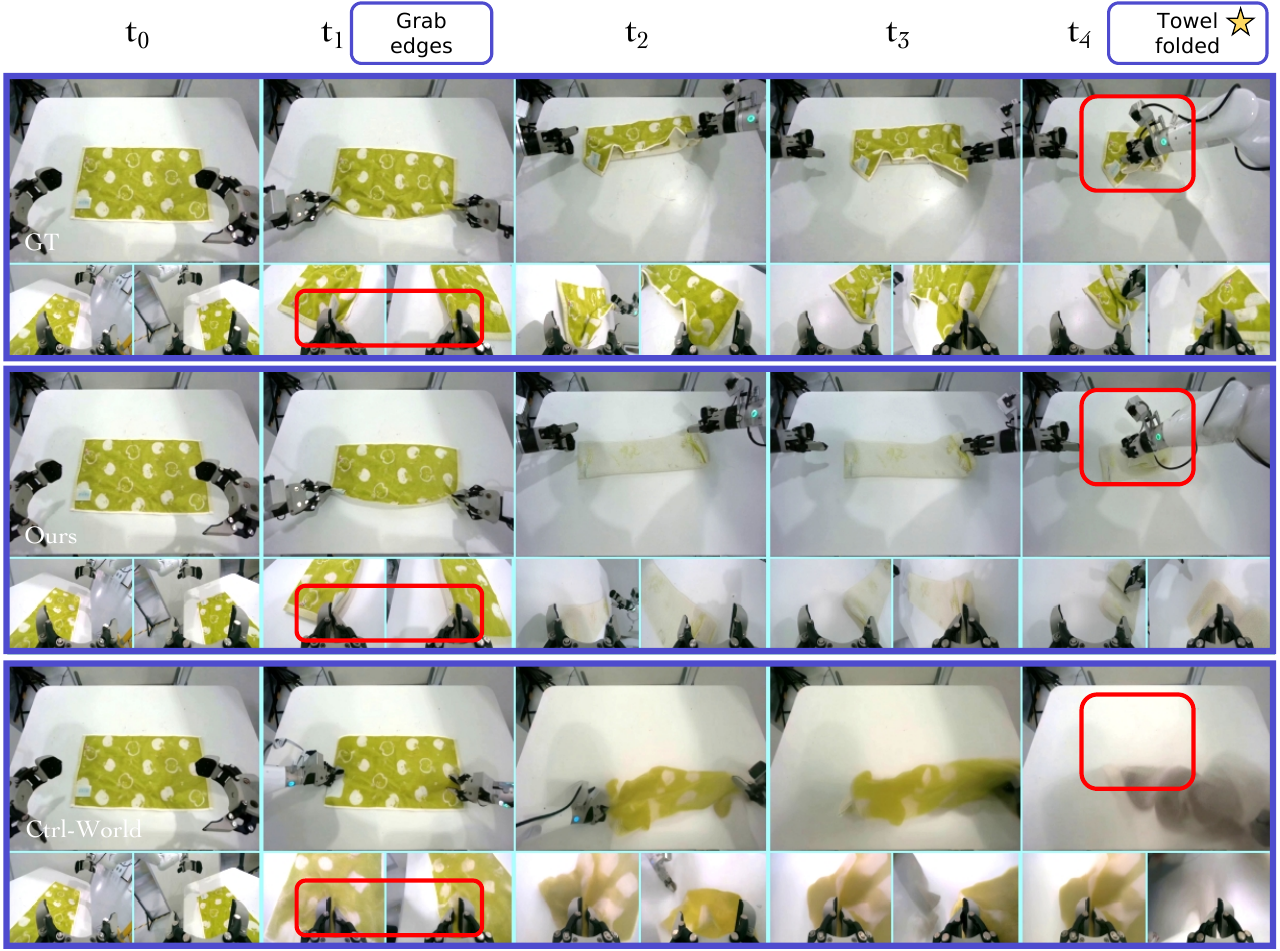}
    \captionof{figure}{Fold towels (Multi-View).}
    \vspace{-3mm}
    \label{fig:more_results_towel_three}
\end{center}

\begin{center}
    \centering
    \includegraphics[width=0.88\linewidth]{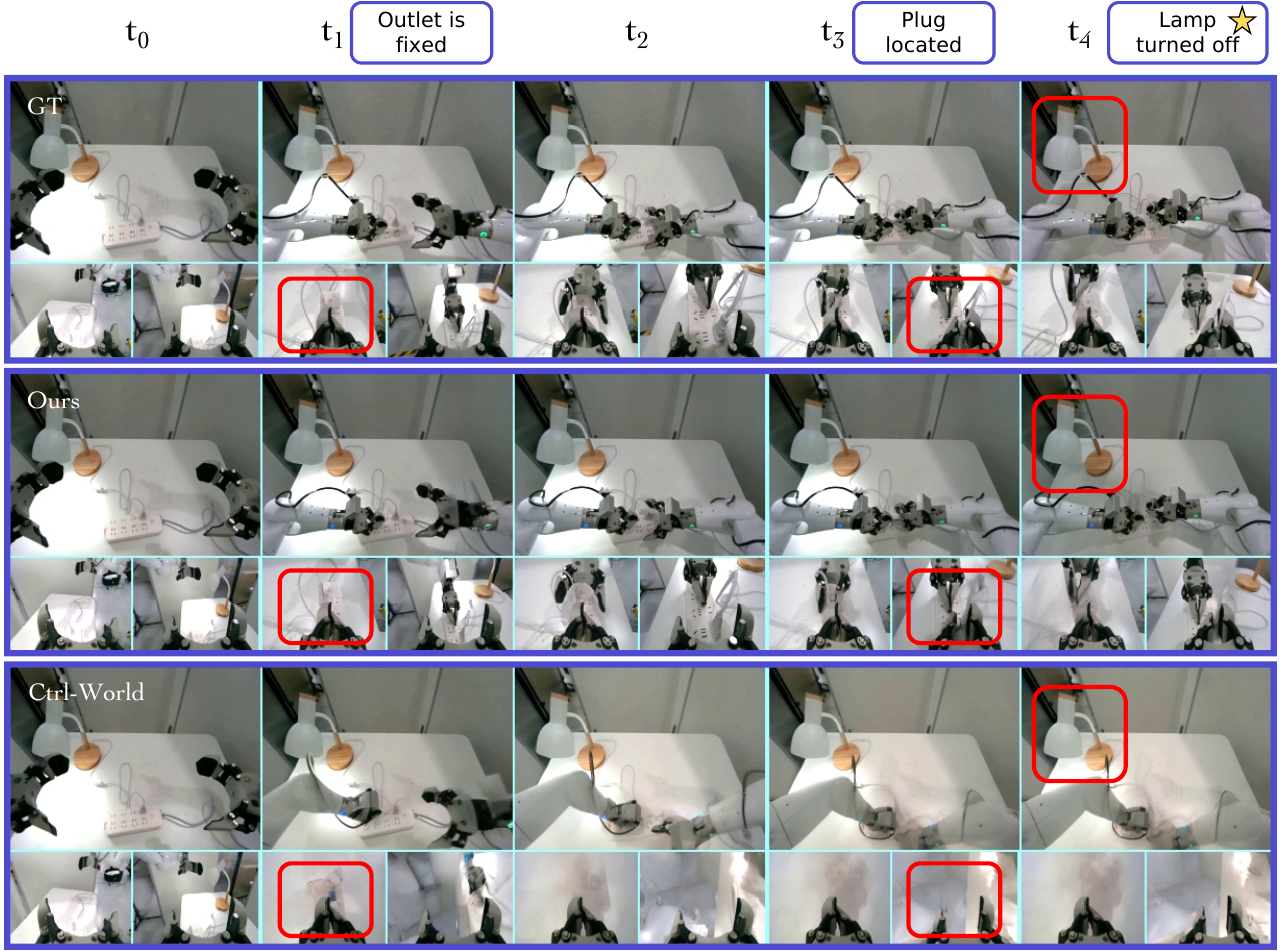}
    \captionof{figure}{Pull out plug (Multi-View).}
    \label{fig:more_results_plug_three}
\end{center}

\begin{center}
    \centering
    \includegraphics[width=0.88\linewidth]{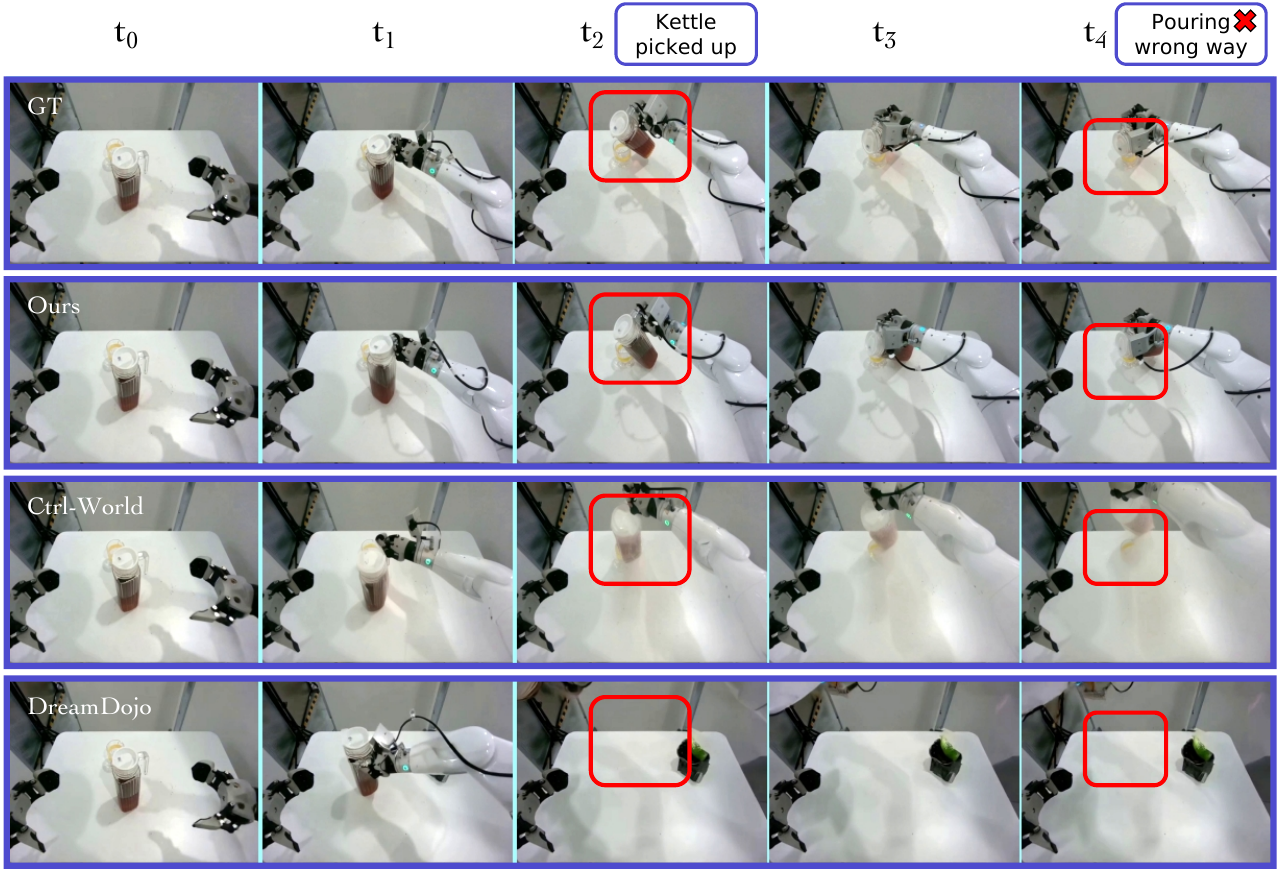}
    \captionof{figure}{Pour water (Head-View).}
    \label{fig:more_results_water_head}
\end{center}

\begin{center}
    \centering
    \includegraphics[width=0.88\linewidth]{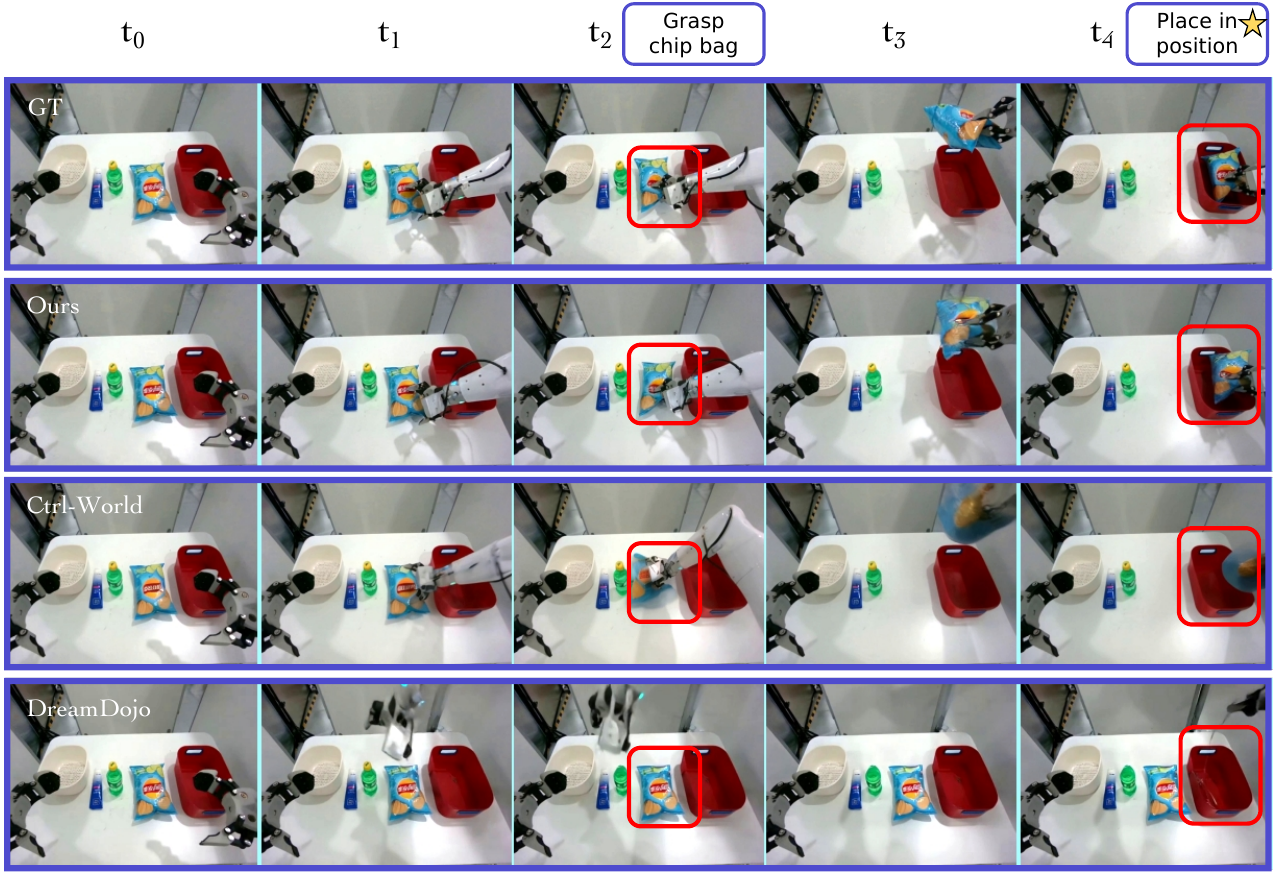}
    \captionof{figure}{Command grasp \& release (Head-View).}
    \label{fig:more_results_grasp_head}
\end{center}

\begin{center}
    \centering
    \includegraphics[width=0.88\linewidth]{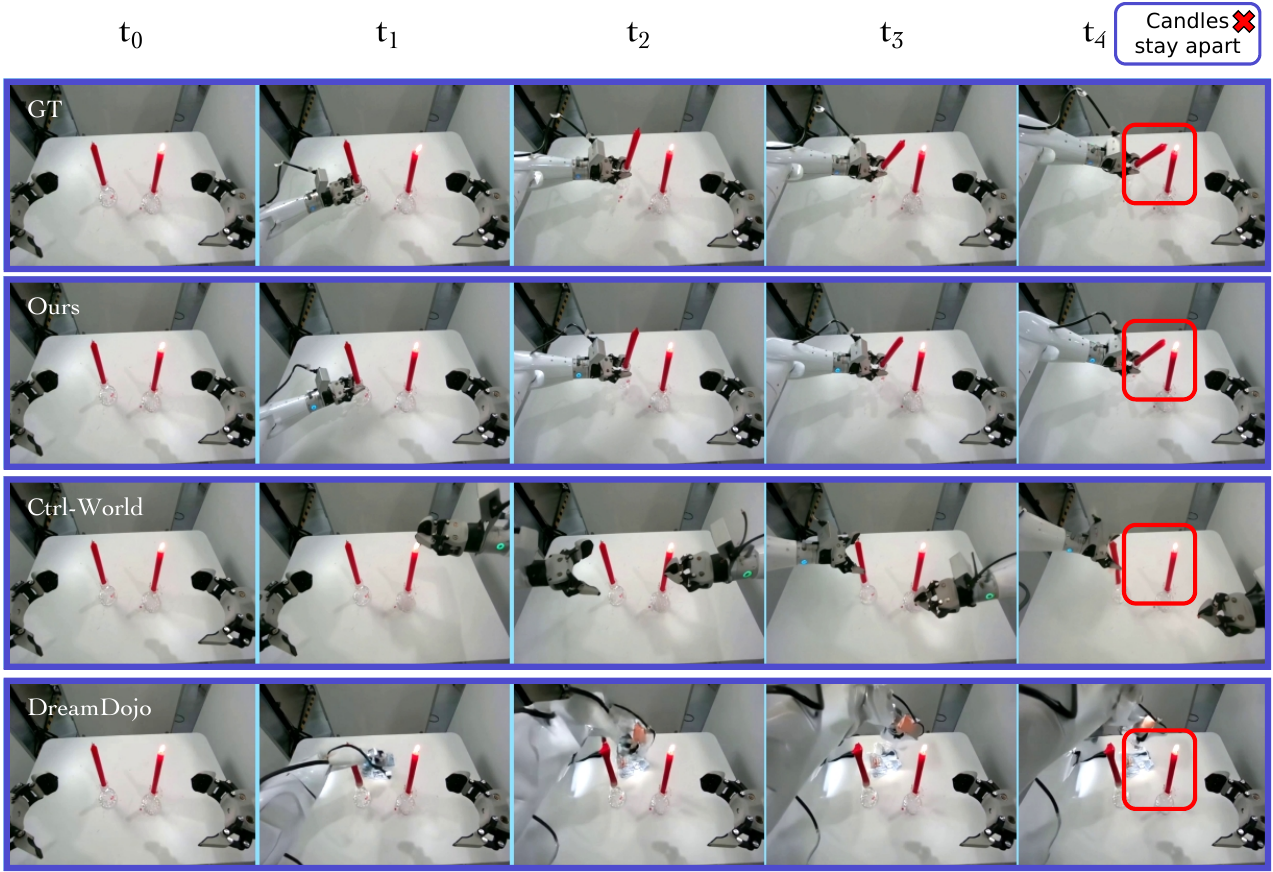}
    \captionof{figure}{Borrow flame (Head-View).}
    \label{fig:more_results_candle_head}
\end{center}

\begin{center}
    \centering
    \includegraphics[width=0.88\linewidth]{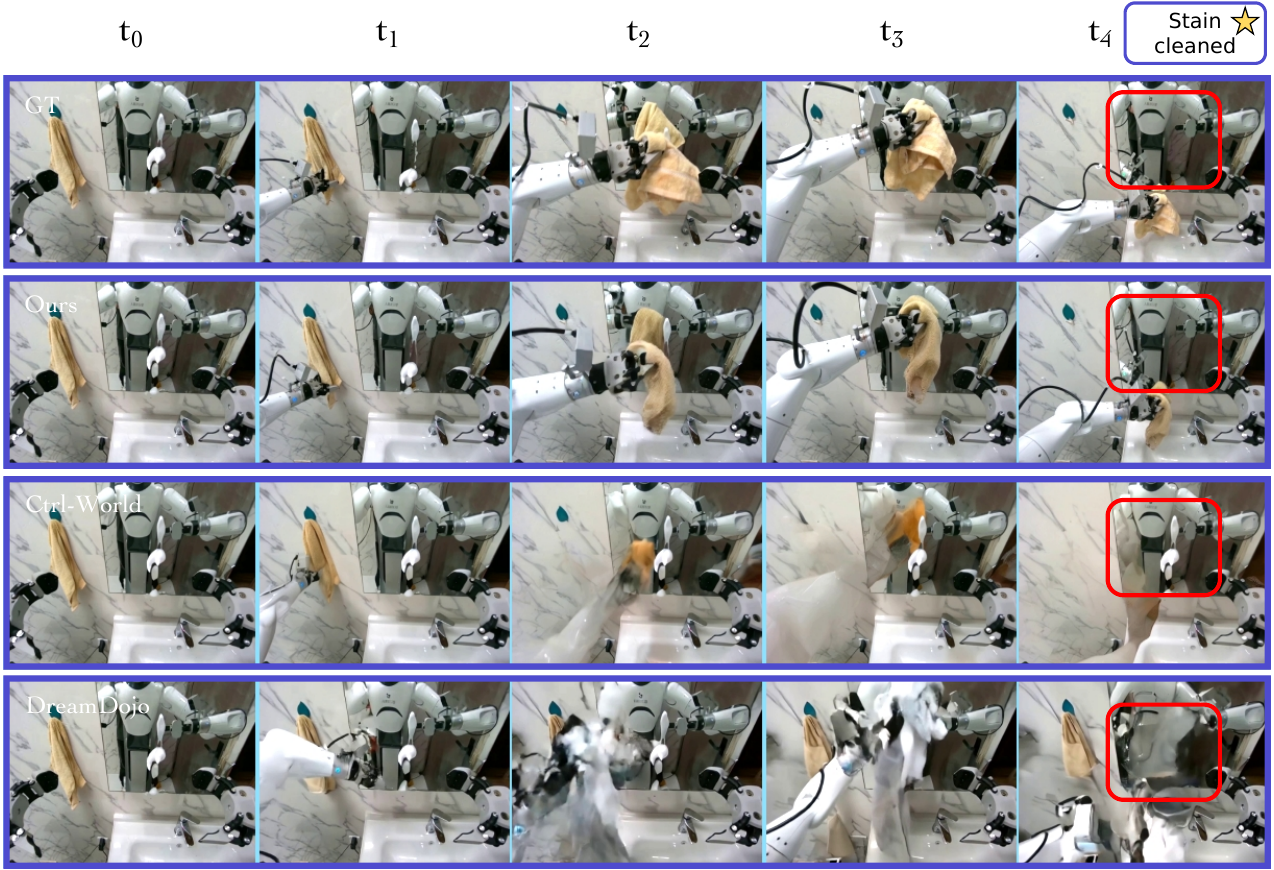}
    \captionof{figure}{Clean mirror (Head-View).}
    \label{fig:more_results_mirror_head}
\end{center}

\begin{center}
    \centering
    \includegraphics[width=0.88\linewidth]{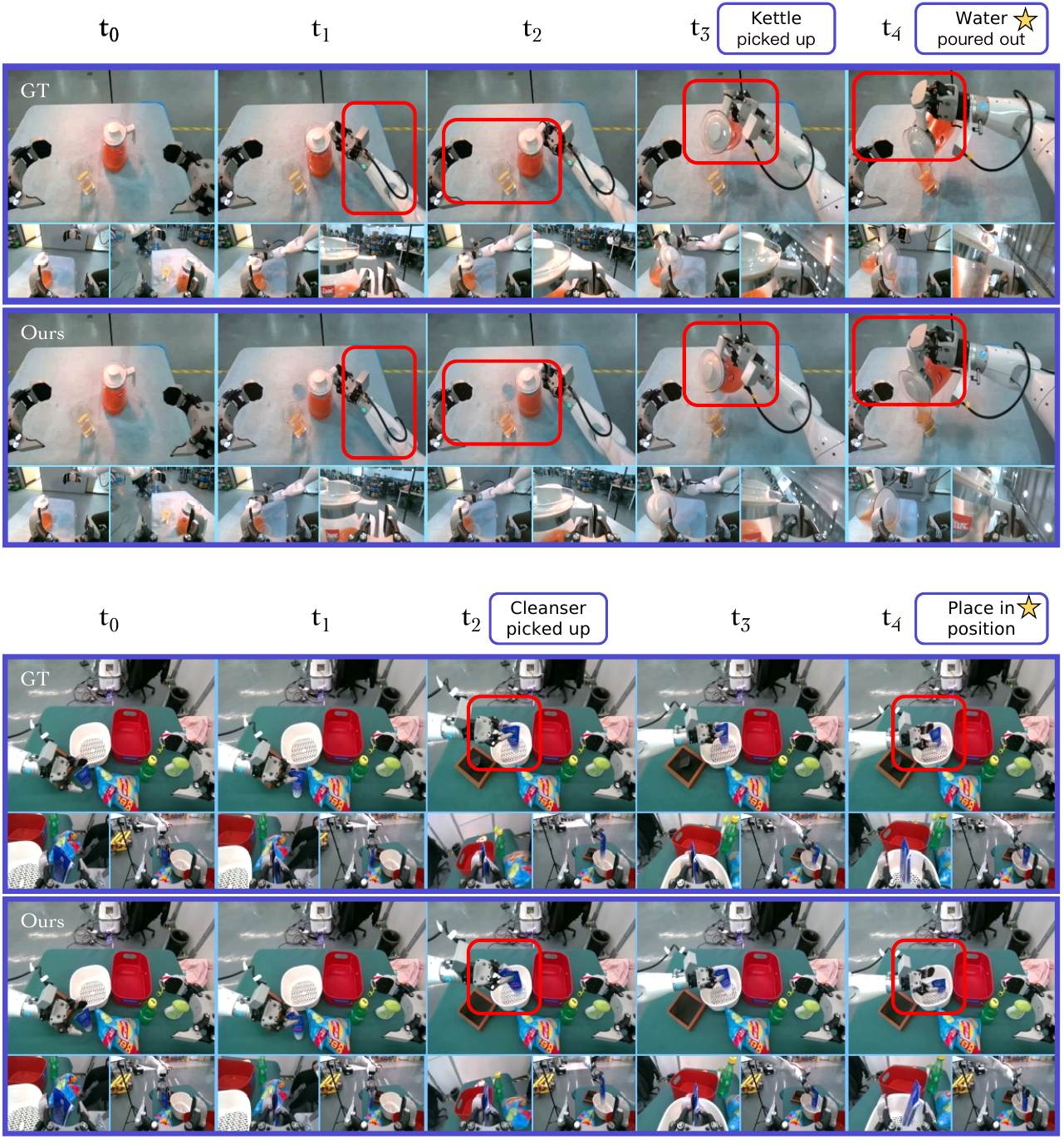}
    \captionof{figure}{More results - GE-Sim 2.0.}
    \label{fig:more_results_our_1}
\end{center}

\begin{center}
    \centering
    \includegraphics[width=0.88\linewidth]{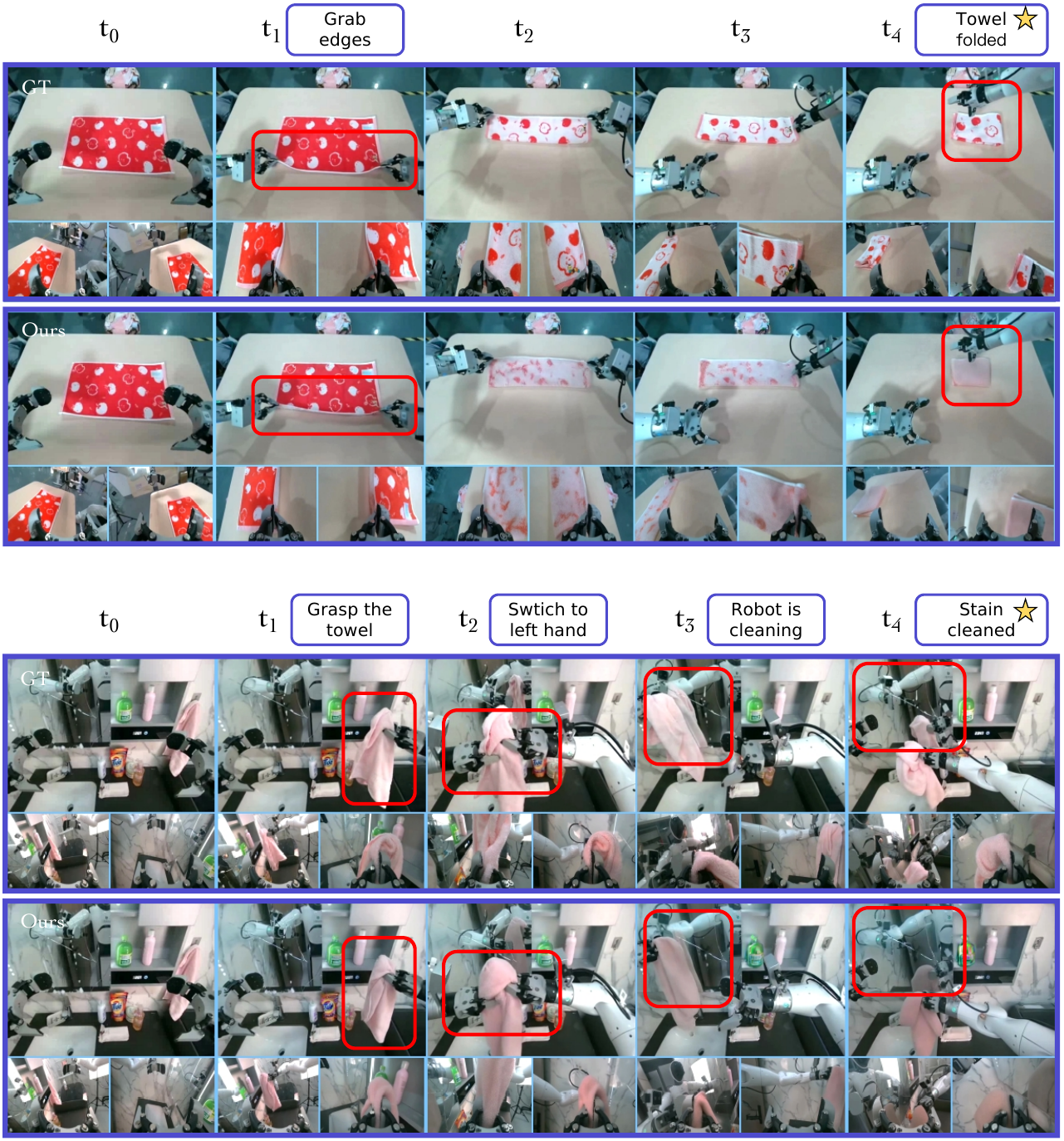}
    \captionof{figure}{More results - GE-Sim 2.0.}
    \label{fig:more_results_our_2}
\end{center}

\end{document}